\documentclass{article}

\usepackage[preprint]{neurips_2026}
\usepackage[utf8]{inputenc}
\usepackage[T1]{fontenc}
\usepackage{hyperref}
\usepackage{url}
\usepackage{booktabs}
\usepackage{amsfonts}
\usepackage{nicefrac}
\usepackage{microtype}
\usepackage{xcolor}
\usepackage{algorithm}
\usepackage{algpseudocode}
\usepackage{amsmath}
\usepackage{graphicx}
\usepackage{enumitem}
\usepackage{multirow}
\usepackage{makecell}
\usepackage[table]{xcolor}
\usepackage{array}
\usepackage{caption}
\usepackage{colortbl}       
\usepackage{subcaption}
\usepackage{wrapfig}
\usepackage[most]{tcolorbox} 
\usepackage{xcolor} 
\usepackage{authblk}
\setlist[itemize]{leftmargin=10pt}

\definecolor{cAbsBlue}{RGB}{25, 95, 205}
\definecolor{cRelGreen}{RGB}{30, 130, 55}
\definecolor{cDeltaBg}{RGB}{230, 241, 255}
\definecolor{cSlash}{gray}{0.45}
\definecolor{cGrpoBg}{gray}{0.91}
\definecolor{cAvgBg}{RGB}{255, 252, 235}

\newcommand{\VS}{\,{\color{cSlash}/}\,}
\newcommand{\scr}[2]{#1\VS#2}
\newcommand{\bscr}[2]{\textbf{#1}\VS\textbf{#2}}

\newcommand{\DC}[4]{%
  \makecell[c]{%
    {\small\color{red}\bfseries +#1\VS+#2}\\[-1pt]%
    {\small\color{cAbsred}\itshape +#3\%\VS+#4\%}%
  }%
}

\renewcommand{\arraystretch}{1.35}
\newcolumntype{C}{>{\centering\arraybackslash}p{2.70cm}}
\newcolumntype{A}{>{\centering\arraybackslash}p{2.38cm}}

\title{Exploiting Verification-Generation Gap:  Test-Time Reinforcement Learning with Confidence-Conditioned Verification}

\usepackage{authblk}

\author[1]{Jiahui Li\textsuperscript{*}}
\author[1]{Jianfeng Shan\textsuperscript{*}}
\author[1]{Wenpei Chen}
\author[1]{Shunyu Wu}
\author[1]{Jian Lou}
\author[2]{Wenjie Feng}
\author[1]{Dan Li}
\author[3]{See-Kiong Ng}
\affil[1]{Sun Yat-Sen University}
\affil[2]{University of Science and Technology of China}
\affil[3]{National University of Singapore }
\affil[ ]{
    \texttt{\{lijh555,shanjf,chenwp28,wusy88\}@mail2.sysu.edu.cn} 
}
\affil[ ]{
    \texttt{\{louj5, lidan263\}@mail.sysu.edu.cn}
}
\date{}

\begin{document}

\maketitle
\begingroup
\renewcommand{\thefootnote}{\fnsymbol{footnote}}
\footnotetext[1]{Equal contribution.}
\endgroup

\begin{abstract}
Test-time reinforcement learning (TTRL) has emerged as a promising paradigm for enhancing the complex reasoning abilities of large language models (LLMs) in a completely label-free manner. Despite existing studies focusing on Pass@1 performance, optimizing Pass@k remains under-explored yet critical in label-free settings, which measures generation coverage for sustained exploration. Optimizing Pass@k in label-free setting is highly non-trivial, as directly applying the Pass@k advantage designs effective for RLVR yields unsatisfactory performance. Through in-depth empirical analysis, we discover the root causes hindering performance: pseudo-label estimations for low-confidence samples have a high probability of being incorrect, while candidate answers for high-confidence samples suffer from severe diversity collapse. To overcome these hurdles, we propose TTRL-CoCoV (\textbf{T}est-\textbf{T}ime \textbf{R}einforcement \textbf{L}earning with \textbf{Co}nfidence-\textbf{Co}nditioned \textbf{V}erification), a novel confidence-adaptive framework that expands Pass@k coverage and improves Pass@1 performance. Based on our key insight that verification capability generally leads generation capability, TTRL-CoCoV employs a confidence-conditioned mechanism: for high-confidence samples, it bootstraps verifier and applies an exploration-enhancing reward to prevent diversity collapse; for low-confidence samples, it delegates pseudo-label selection to the verifier to filter incorrect pseudo-labels; and for medium-confidence samples, it bypasses verification entirely. Extensive experiments demonstrate that TTRL-CoCoV outperforms the best competing methods across 6 widely-recognized benchmarks, achieves average absolute gains of +9.8\% in Pass@1 and +18.7\% in Pass@16 over TTRL, and even achieves absolute Pass@1 improvements of up to +5.0\% across multiple reasoning benchmarks when compared against fully supervised RL methods. Our code repository: \url{https://github.com/shanjf666/CoCoV}.

\end{abstract}

\vspace{-1em}
\section{Introduction}
\label{intro}
Test-time reinforcement learning (TTRL)\citep{zuo2025ttrl,liu2025ettrl,wang2025self,zhou2025evolving} has emerged as a promising paradigm to enhance the reasoning capabilities of large language models (LLMs) in completely label-free settings. Compared to traditional Reinforcement Learning from Verifiable Rewards (RLVR)\citep{jaech2024openai}\citep{guo2025deepseek}, which fundamentally relies on human annotations or oracle-provided labels, TTRL optimizes reasoning policies entirely through self-sampling and internal reward construction. Building upon standard architectures such as Group Relative Policy Optimization (GRPO)\citep{shao2024deepseekmath}, recent TTRL methods effectively address the scalability bottlenecks of RLVR and achieve single-pass generation (Pass@1) performance that closely approaches fully supervised baselines. Beyond Pass@1 accuracy, maintaining a broad exploratory space is critical to prevent premature convergence during policy optimization. In the RLVR setting, recent studies \citep{chen2025pass,walder2025pass} have demonstrated that explicitly optimizing for Pass@k generation coverage effectively balances exploration and exploitation. However, despite its proven importance in supervised environments, optimizing Pass@k generation coverage remains largely under-explored within the TTRL setting.

To advance the efficacy of Test-Time Reinforcement Learning (TTRL)\citep{zuo2025ttrl}, current methodologies mainly focus on establishing the reliability of self-generated pseudo-labels. For complex, low-consistency problems, recent sample-level advancements successfully employ consensus gating and confidence reweighting to secure label purity and stabilize the optimization process \citep{yan2026if,yu2025restrain}. Alternatively, other approaches leverage internal self-verification mechanisms or an outer verifier to actively filter out spurious trajectories or perform autonomous error correction\citep{pan2026coverrl,yan2026if,liao2026tool}. Simultaneously, researchers have also recognized the importance of preventing premature convergence—often observed as a dramatic collapse in the standard deviation of response lengths on mastered problems \citep{chen2025pass,walder2025pass}. To address this, recent studies have thoughtfully introduced novelty or entropy bonuses to maintain a broad exploratory space in unsupervised TTRL\citep{liu2025ettrl,zhou2025evolving}. Collectively, these pioneering efforts have advanced TTRL along two directions: managing label reliability and enhancing exploration. Yet, a fundamental dimension remains surprisingly unexplored: in completely unsupervised TTRL, how to systematically optimize Pass@k generation coverage, which is a critical metric for sustained exploration. This gap is non-trivial, as we will show through three empirical insights below.

Directly extending standard Pass@k optimization to the label-free TTRL environment faces three fundamental challenges, which we identify through systematic empirical analysis. \textbf{First}, the challenge of noisy exploration signals: Pass@K optimization inherently relies on exploring low-confidence samples to broaden the exploratory space \citep{chen2025pass}. However, in completely unsupervised TTRL, this introduces severe pseudo-label noise that corrupts the required exploration signals in Pass@K optimization.  \textbf{Second}, the challenge of length collapse: While Pass@k successfully sustains response diversity in supervised RLVR, our analysis reveals that applying it in TTRL fails to prevent a severe collapse in response length standard deviation, which acts as a latent leading indicator of premature accuracy stagnation. \textbf{Third}, the challenge of empirical formulation failure: Consequently, directly adopting standard supervised Pass@k advantage formulations fails to expand the exploratory space in label-free settings, causing actual Pass@k performance to inevitably degrade over time. (A comprehensive evaluation and proof of these empirical challenges are provided in Section  \ref{motivation}). 

Motivated by these empirical findings, we propose TTRL-CoCoV (\textbf{T}est-\textbf{T}ime \textbf{R}einforcement \textbf{L}earning with \textbf{Co}nfidence-\textbf{Co}nditioned \textbf{V}erification), a novel framework featuring targeted mechanisms to explicitly resolve each identified challenge. Corresponding to the challenge of noisy exploration signals, we introduce a Confidence-Conditioned Classifying mechanism. By adaptively allocating internal verification resources based on per-sample confidence, it drives co-evolution under a unified optimization objective: bootstrapping the verifier on high-confidence successes to actively filter spurious pseudo-labels on low-confidence hard problems, while safely bypassing medium-confidence boundaries. To overcome the persistent trajectory collapse and the empirical failure of naive Pass@k adaptations, we design an exploration-enhancing reward mechanism ($R_{div}$). This tailored mechanism explicitly incentivizes solution-strategy diversity exclusively on high-confidence problems, successfully adapting the core benefits of Pass@k optimization to the label-free setting without corrupting the pseudo-label pool. 

We summarize our primary contributions as follows: 
\begin{itemize} \item \textbf{Empirical Diagnosis of Label-Free Pass@k Dynamics:} We are the first to systematically investigate the under-explored gap of Pass@k generation optimization within completely unsupervised TTRL. Through rigorous empirical analysis, we identify the fundamental bottlenecks, including noisy exploration signals, latent diversity collapse, and formulation mismatches that make this adaptation highly non-trivial in label-free environments.
\item \textbf{A Novel Confidence-Adaptive Framework (TTRL-CoCoV):} To overcome these challenges, we propose TTRL-CoCoV. It employs a Confidence-Conditioned Classifying mechanism that drives co-evolution between the generator and verifier under a unified optimization objective, effectively filtering noisy exploration signals. Building upon this purified signal, we successfully adapt Pass@k optimization to the unsupervised setting via a tailored exploration-enhancing reward, safely sustaining broad exploratory exploration without corrupting the pseudo-label pool.

\item \textbf{Comprehensive Evaluations and SOTA Performance:} We conduct exhaustive experiments across six complex reasoning benchmarks (e.g., MATH500, AIME24/25) using multiple backbone models (Qwen3-4B/8B). Empirically, TTRL-CoCoV completely reverses the unsupervised Pass@k degradation, yielding massive exploratory generation coverage (an average absolute gain of +18.7\% in Pass@16). Furthermore, via a two-stage annealing strategy, our fully unsupervised method achieves absolute Pass@1 improvements of up to +5.0\%, successfully breaking the performance ceiling established by fully supervised GRPO baselines.
\end{itemize}

\section{Backgrounds and Preliminary}
\label{gen_inst}
\textbf{Problem Setup.} Let $\mathcal{D}$ be a dataset of reasoning questions. For a given prompt $\boldsymbol{x} \sim \mathcal{D}$, a target large language model (LLM) policy $\pi_\theta$ generates a response sequence $\boldsymbol{y} = (y_1, \ldots, y_{|\boldsymbol{y}|})$ autoregressively. Specifically, $\pi_\theta(y_t \mid \boldsymbol{x}, \boldsymbol{y}_{<t})$ denotes the probability of generating the $t$-th token $y_t$ conditioned on the prompt $\boldsymbol{x}$ and prefix $\boldsymbol{y}_{<t}$. During reinforcement learning, responses are sampled via the old policy $\pi_{\theta_{\text{old}}}$ and evaluated against the ground-truth label $y^{\text{gt}}$ to yield a reward $R$, guiding the policy optimization from a pre-trained reference model $\pi_{\text{ref}}$.

\textbf{Group Relative Policy Optimization (GRPO).} To eliminate the independent value network, GRPO \citep{shao2024deepseekmath} estimates the advantage by normalizing intra-group relative rewards. For a given $\boldsymbol{x}$, the policy samples $N$ responses $\{\boldsymbol{y}_i\}_{i=1}^N$. Each receives a reward $R_i \in \{R_{\text{neg}}, R_{\text{pos}}\}$ corresponding to an incorrect or correct response based on $y^{\text{gt}}$. The sequence-level advantage $\hat{A}_i$ is then calculated by standardizing these rewards within the sampled group. This relative advantage is subsequently utilized to optimize the policy via a standard clipped surrogate objective. The complete mathematical formulations for the GRPO advantage estimation and policy loss are provided in Appendix \ref{grpo}.

\textbf{Pass@k Training.} Recent works \citep{chen2025pass, walder2025pass, wan2026dsdr} directly optimize the Pass@k metric, defined as the probability of generating at least one correct answer within $k$ samples. Specifically, Pass@k Training \citep{chen2025pass} derives analytical advantages, denoted as $A^{pass@k}$, based on the combinatorial probabilities of sampling correct versus incorrect responses within a group. While these closed-form formulations (detailed in Appendix \ref{passk}) effectively encourage exploration, they rely strictly on ground-truth labels to accurately identify positive and negative responses. Extending such Pass@k optimization objectives to the completely label-free Test-Time Reinforcement Learning \citep{zuo2025ttrl} setting remains under-explored.

\section{Motivation}
\label{motivation}
To guide our framework, we conduct controlled empirical analyses on TTRL training dynamics. These reveal three empirical insights that diagnose existing failure modes and directly motivate our algorithmic design.

\subsection{Insight I: Naively Applying Pass@k to TTRL Fails to Sustain Exploration}
The fundamental advantage of supervised Pass@k training lies in maximizing the utilization of exploratory signals from low-consistency samples to broaden the exploratory space \citep{chen2025pass}. However, empirical analysis reveals that naively adapting the Pass@k objective to TTRL only partially recovers the severe Pass@k collapse triggered by standard label-free training (detailed experimental setups and full results are provided in Appendix \ref{insigt1}). This failure stems from an objective formulation mismatch: unlike supervised RL where ground-truth labels ensure reliability, self-generated exploratory signals in label-free settings are inherently plagued by severe pseudo-label noise. Naively rewarding these unreliable,low-confidence signals degrades rather than expands actual Pass@k performance over time. This establishes a fundamental design requirement: \textbf{to safely harness Pass@k's exploration benefits in TTRL, the framework should first resolve pseudo-label noise in low-consistency samples, which is a prerequisite that naive objective adaptation entirely ignores.} We next investigate whether internal self-verification can serve this role.

\subsection{Insight II: Verification Outperforms Generation Across Most Confidence Levels with Variable Margins}

\label{insight2}
\begin{figure}[htbp]
  \centering
  
  \begin{subfigure}[b]{0.45\linewidth}
    \centering
    \includegraphics[width=0.85\linewidth]{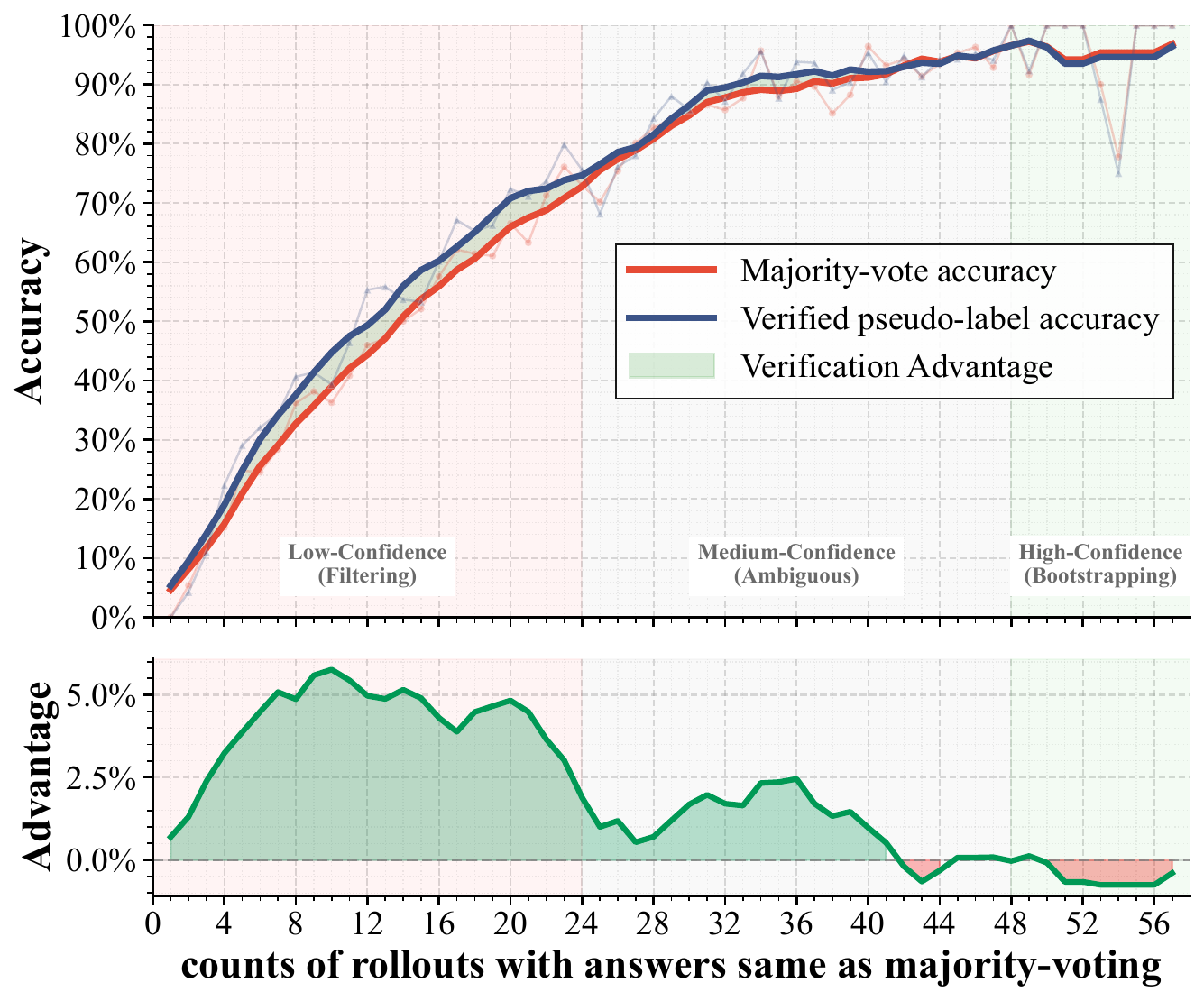}
    \caption{}
  \end{subfigure}
  \hfill 
  \begin{subfigure}[b]{0.45\linewidth}
    \centering
    \includegraphics[width=0.9\linewidth]{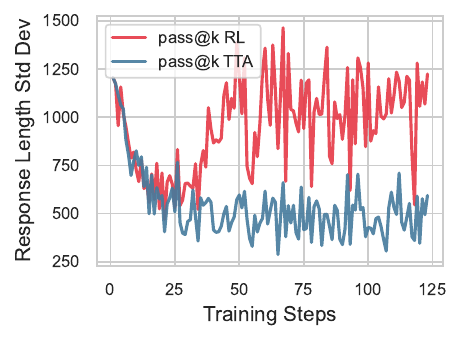}
    \caption{}
  \end{subfigure}
  
  \vspace{-0.8em} 
  
  \caption{Key empirical insights motivating our framework. (a) Asymmetric Verification Advantage: The verifier effectively filters low-confidence noise but provides near-zero gain on high-confidence samples, necessitating a \textit{confidence-conditioned} strategy. (b) Length Variance Collapse: Pass@k TTA suffers a severe length variance collapse due to over-fitting on "safe" majorities, dictating that diversity incentives must be injected \textit{exclusively} into high-confidence problems.}
  \label{fig:training_dynamics}

\end{figure}

To address pseudo-label reliability more directly, we introduce an internal verifier role, tasking the model with evaluating its own outputs to refine the training signal. As an observation, we find that the model's verification capability generally leads its generation capability. However, we also discover a critical asymmetry: the verifier's performance is not uniform but varies significantly with the generator's confidence level. Mapping the verification advantage (the accuracy gain of verified pseudo-labels over standard majority voting) across the confidence spectrum (Figure~\ref{fig:training_dynamics}a) reveals a clear three-regime structure. In the low-confidence regime, the verification advantage is substantial (peaking above 5\%), as the verifier reliably filters spurious candidates the generator cannot yet produce correctly. In the medium-confidence regime, the advantage shrinks to near zero, as the verifier is roughly as uncertain as the generator, offering no exploitable advantage. In the high-confidence regime, the advantage vanishes entirely, as majority voting is already reliable. These samples can serve a distinct role: because their pseudo-labels are near-certain, they provide clean signal for continuously training and improving the verifier's discriminative capability, which in turn strengthens its filtering performance on low-confidence samples in future iterations. The implication is clear: \textbf{effective pseudo-label refinement in TTRL must be confidence-conditioned, not uniform}. This co-evolution, where the generator's mastered problems train the verifier, which in turn guides the generator through uncertain regimes, forms the conceptual backbone of our framework and motivates our confidence-conditioned verification strategy.

\vspace{-1.5em}

\subsection{Insight III: Length Standard Deviation Collapse Signals Exploration Failure}
\label{insight3}
To understand the mechanistic cause of the Pass@k collapse observed in Insight I, we track response length standard deviation (Figure~\ref{fig:training_dynamics}b). While Pass@k RL (oracle) and Pass@k TTA maintain similar variance early on, a sharp divergence emerges in mid-training. Pass@k RL sustains high variance (800–1,500 tokens) reflecting active exploration, whereas Pass@k TTA steadily collapses to a narrow plateau (400–650 tokens). This variance collapse acts as a leading indicator, temporally preceding the Pass@k accuracy drop. We attribute this collapse to the progressive over-optimization of high-consensus samples: as training repeatedly reinforces these "safe" majority paths, initially diverse reasoning strategies converge into rigid, repetitive templates, while the absence of reliable rewards for novel trajectories leaves the model with no incentive to explore alternatives. The persistent gap in length variance between the two settings mirrors and likely underlies the corresponding gap in Pass@k accuracy, suggesting that \textbf{response length diversity is a concrete and measurable metric for exploration health in TTRL.} Accordingly, an effective TTRL framework should explicitly inject exploration incentives on high-confidence problems where pseudo-label correctness is already guaranteed without corrupting the signal on uncertain samples.

\section{Proposed Method}
\label{headings}
Guided by the empirical insights in Section~\ref{motivation}, we introduce the TTRL-CoCoV framework, designed to simultaneously optimize label purity and generation diversity in a completely label-free setting. At its core, our approach utilizes a single shared-weight model $\pi_\theta$ that fulfills the dual functional roles of generator and verifier. By dynamically classifying samples according to generation confidence, the framework drives co-evolution between these two roles to effectively filter noisy pseudo-labels. Coupled with a tailored exploration-enhancing reward to prevent trajectory collapse, TTRL-CoCoV successfully adapts Pass@k optimization to TTRL under a unified optimization objective.
\begin{figure}
  \centering
  \includegraphics[width=1\linewidth]{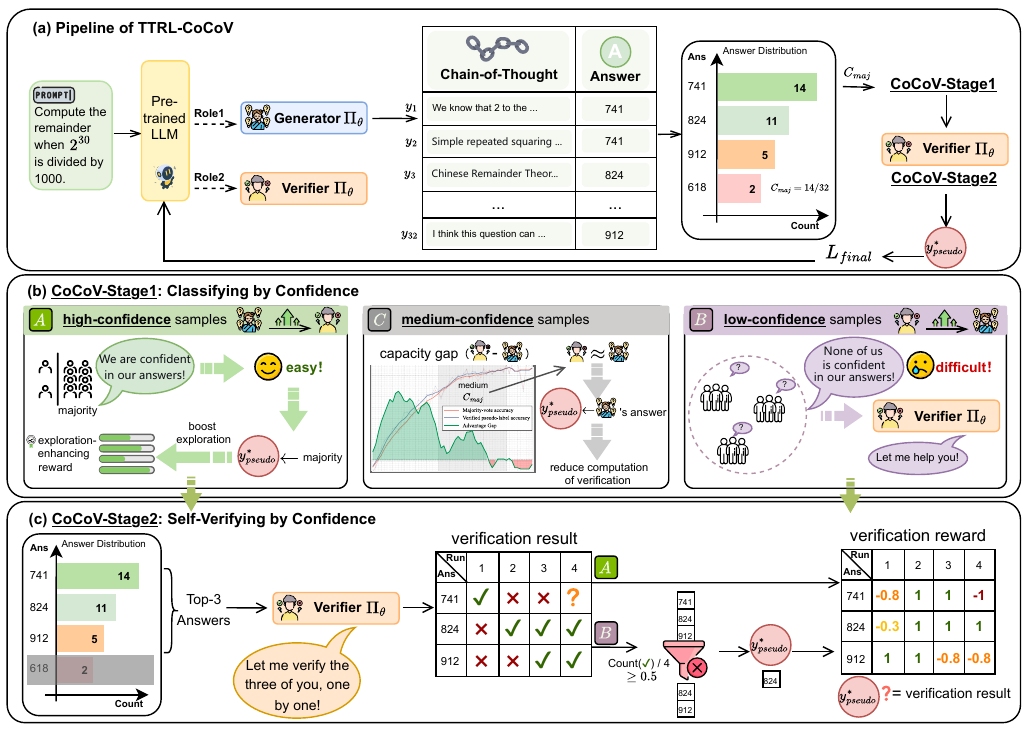}

  \caption{\textbf{(a) Overview of TTRL-CoCoV}, which employs a shared-weight model ($\pi_\theta$) as both generator and verifier to sample trajectories and establish answer consensus. \textbf{(b) CoCoV-Stage 1 (Classifying by Confidence):} Based on consensus confidence, high-confidence samples receive an exploration-enhancing reward and activate the verifier for training; low-confidence samples wait for explicit verification; medium-confidence samples bypass the verification stage and update the generator with majority-labels. \textbf{(c) CoCoV-Stage 2 (Self-Verifying by Confidence):} For high-confidence samples, majority-labels are used to update the verifier; for low-confidence samples, the verifier selects refined pseudo-labels to update both the generator and the verifier.}
  \label{fig:example}

\end{figure}

\subsection{Overall Pipeline}
The overall training process dynamically balances generation, verification and exploration based on the model's internal confidence. Given a question $x$, the model $\pi_\theta$ first operates in generation mode to sample $N$ reasoning trajectories and extract Top-$K$ candidate answer set. We compute the proportion of the majority answer within the current batch and define it as the majority confidence $C_{maj}$. Based on $C_{maj}$ and the preset thresholds ($\tau_{low}$, $\tau_{high}$), each sample is classified into one of three training optimization zones (detailed in Section~\ref{sec:classification}). If the zone triggers the verification mechanism, the model switches to the verifier role and generates $m$ independent verification trajectories for each Top-$K$ candidate answer, thereby determining the final pseudo-label $y_{pseudo}^*$ and its corresponding confidence weight $\mathcal{C}_{maj}$. Finally, the overall optimization objective is a dynamically weighted sum of the first-stage generation policy loss $\mathcal{L}_{first}$ and the second-stage verification policy loss $\mathcal{L}_{second}$:
\begin{equation}
\small
\mathcal{L}_{total} = \mathcal{L}_{first} + \mathbb{I}_{verify} \cdot \mathcal{C}_{maj} \cdot \left( \frac{1}{K} \sum_{j=1}^{K} \mathcal{L}_{second}^{(j)} \right).
\end{equation}
where $\mathbb{I}_{verify} \in \{0, 1\}$ is an indicator controlling whether the verification stage is activated, and $j$ indexes the $j$-th candidate answer. $\mathcal{C}_{maj}$ serves as a dynamic weight to prevent low-consistency, high-noise samples from dominating the overall gradient. Averaging the verification loss over $K$ candidate answers further stabilizes the direction of the verifier update. Algorithm~\ref{alg:smd_ttrl} outlines the complete training procedure.

\subsection{TTRL-CoCoV Stage1: Conditioning By Confidence}
\label{sec:classification}
Based on $C_{maj}$ and thresholds $\tau_{low}$ and $\tau_{high}$, each sample is classified into one of three regions.

\subsubsection{Region A: High-Confidence Exploration Zone}
\label{sec:A}
When $C_{maj} \ge \tau_{high}$, the model exhibits extremely high consensus on a problem and it can be considered sufficiently confident in its solutions towards this problem. Therefore,  we directly designate the majority answer $y_{maj}$ as the final pseudo-label $y_{pseudo}^*$ to compute the generation policy loss. Given the high accuracy of the pseudo-label in this region, we incorporate two designs to better leverage it.

\paragraph{Verifier Optimization}
Although the model already possesses high confidence in the generation stage, we still activate verification ($\mathbb{I}_{verify}=1$). The core motivation is not screening or correction, but rather providing high-quality training data for the verifier, thereby improving its verification capability and enabling co-evolution of the generator and verifier.

\paragraph{Exploration-Enhance Reward}
Motivated by \hyperref[insight3]{Insight III}, which identifies the collapse of length standard deviation as a leading indicator of premature convergence, we aim to actively sustain trajectory diversity on mastered problems. Since longer trajectories often contain more detailed reasoning steps while shorter ones represent shortcuts to the answer, both are worth rewarding to maintain solution diversity. Therefore, we introduce a length diversity reward for correctly labeled samples in the high-consistency region to stimulate exploration:

\begin{equation}
\small
R_{div}(y_i) = \lambda \cdot \min\left( \frac{|l_i - \mu_L|}{\sigma_L + \epsilon},\ C_{max} \right).
\end{equation}
where $l_i$ is the token length of the current trajectory, $\mu_L$ and $\sigma_L$ are the mean and standard deviation of positive sample lengths in the current batch. We set $\lambda = 0.05$ and $C_{max} = 2$. The final advantage is computed as:
\begin{equation}
\small
\tilde{A}_i = A_i^{pass@k} + \mathbb{I}[r_i = 1] \cdot R_{div}(y_i).
\end{equation}
To prevent positive advantages from becoming negative after normalization under extreme sample distributions, we apply an advantage-clipping mechanism:
\begin{equation}
\small
A_i^{\text{final}} = 
\begin{cases} 
\max\left(\tilde{A}_i,\ \operatorname{Norm}(\tilde{A}_i)\right), & r_i = 1. \\ 
\operatorname{Norm}(\tilde{A}_i), & r_i = 0.
\end{cases}
\end{equation}
where $\mathrm{Norm}(\cdot)$ denotes within-group mean-variance normalization.

\subsubsection{Region B: Low-Confidence Verification Zone}
When $C_{maj} < \tau_{low}$, the sample falls within the low-confidence region where the generator exhibits low confidence in its answers to the problem. The accuracy of the pseudo-labels produced by majority voting is low, which inevitably introduces significant pseudo-label noise and, in turn, corrupts the required exploration signals in pass@k optimization. To mitigate this risk and based on the previous Insight II that verification capacity is generally stronger than generation, we rely on the verifier as a funnel to filter out erroneous answers with relatively high confidence. We set $\mathbb{I}_{verify}=1$ and feed the Top-$K$ candidate answers into the self-verification stage. If the verifier successfully identifies high-confidence answers from the candidate set, we designate the best candidate as the final pseudo-label $y_{pseudo}^*$ and perform joint updating. If no candidate passes verification, we skip gradient updates for this sample to avoid contaminating the model with erroneous gradients.

\subsubsection{Region C: Medium-Confidence Transition Zone}
When $\tau_{low} \le C_{maj} < \tau_{high}$, the sample lies within the moderate confidence region, where the majority answer and the answer filtered by the verifier are essentially equivalent in terms of accuracy. Within this range, the generator lacks both the overwhelming consensus required to provide high-quality positive examples and the confidence necessary to conduct extensive edge exploration based on the assumption that the pseudo-label is correct. Therefore, we skip the self-verification stage entirely ($\mathbb{I}_{verify} = 0$) to save inference overhead and avoid introducing worse noise signals. The samples in this zone use only the majority answer $y_{maj}$ generated in the first stage as a pseudo-label for generator updates, without applying any additional exploration reward signals.

\subsection{TTRL-CoCoV Stage2: \text{Self-Verifying} based on Confidence Conditioning}
When self-verification is triggered, the model $\pi_\theta$ verifies the extracted Top-$K$ candidate answers $\{o_k\}_{k=1}^{K}$ to assess the validity of each and filter out obviously incorrect ones. To quantify the verification results, we introduce the Verification Pass Rate (VPR). For a candidate answer $a_k$, the model generates $m$ verification trajectories, and its VPR is defined as the proportion of trajectories that output True:
$
\small
\text{VPR}(o_k) = \frac{1}{m}\sum_{j=1}^{m} \mathbb{I}[\text{verify}^{(j)}(o_k) = \text{True}].
$ The computation of the verifier reward $R_{\text{second}}^{(j,k)}$ depends heavily on the classification region, as the source of the supervisory pseudo-label differs across regions:

\textbf{Region A}: Since $y_{maj}$ is highly reliable, it is directly used as the pseudo-label ($y_{pseudo}^* = y_{maj}$), and the verifier's predictions are compared against this pseudo-label to compute rewards. \\
$\textbullet$ \textbf{Region B}: To filter the severe pseudo-label noise inherent in uncertain regimes, only candidate answers with $\text{VPR}(o_k) > 0.5$ are admitted into the trusted set $\mathcal{A}_{true}$. We then select the answer with the highest first-stage consensus from $\mathcal{A}_{true}$ as the pseudo-label. The verifier's reward is computed based on whether its prediction matches this dynamically mined label. If $\mathcal{A}_{true} = \emptyset$, we assume that none of the Top-$K$ candidates is likely to be correct and skip the update for this sample.

When computing the verifier's policy loss $\mathcal{L}_{second}$, we adopt the principle of ``lenient to false negatives while strict with false positives''. The primary function of the verifier is not to directly select the correct answer, but to intercept potentially erroneous reasoning that may hide behind high confidence. We design an asymmetric soft reward matrix for verification trajectories (the precise reward formulation is detailed in Appendix \ref{reward}). By assigning a heavier penalty to false positives (FP), the verifier is shaped into a more stringent screening mechanism. The above rewards are normalized at the prompt level to obtain the final verifier loss $\mathcal{L}_{second}$, which is added to the first-stage policy loss $\mathcal{L}_{first}$ to update $\theta$, forming a closed training loop of generator exploration and verifier filtering.

\section{Experiments}
\label{experiments}

\subsection{Set up}
Detailed experiment setup (model families, datasets, and evaluation metrics) is in Appendix \ref{setup}. 

\paragraph{Baselines}
We compare with three existing test-time reinforcement learning methods. (1) \textbf{TTRL} \citep{zuo2025ttrl}, which samples multiple trajectories from unlabeled data and leverages the majority vote as pseudo-labels to formulate reward signals; (2) \textbf{SCRL} \citep{yan2026if}, a consensus-driven approach that reinforces correct reasoning paths with positive pseudo-labels when strong agreement is reached, while using generative uncertainty to construct negative pseudo-labels for pruning incorrect paths otherwise; and (3) \textbf{Co-Rewarding-III} \citep{zhang2025co}, which enhances reward stability via multi-view cross-validation. The latter assigns positive rewards only if the model yields consistent answers across semantically rewritten problem variants, while additionally employing an Exponential Moving Average (EMA) teacher to provide reference pseudo-labels for policy updates.

\definecolor{cAbsred}{RGB}{200,30,30}
\definecolor{cSlash}{gray}{0.45}
\definecolor{cGroupBg}{RGB}{235,235,235}

\setlength{\tabcolsep}{5pt}
 
\begin{table}[t]
\centering

\renewcommand{\arraystretch}{1.3} 
\caption{
  Main experimental results on six reasoning benchmarks.
  Each cell reports \emph{pass@1\,/\,pass@16}.
  \textbf{Ours} refers to our proposed TTRL-CoCoV.
  $\Delta$ rows show gains of Ours over TTRL:
  {\color{red}\textbf{absolute}} (top) and
  {\color{cAbsred}\textit{relative\,\%}} (bottom).
  $^\dagger$Uses labeled reward data; all other methods are \emph{label-free}.
}
\label{tab:main}
\small

\resizebox{\linewidth}{!}{
\begin{tabular}{@{\hskip\tabcolsep} l *{6}{>{\centering\arraybackslash}p{2.2cm}} >{\centering\arraybackslash}p{2.2cm} @{\hskip\tabcolsep}}
\toprule
\textbf{Method}
  & \textbf{AIME24} & \textbf{MATH500} & \textbf{AIME25}
  & \textbf{AMC}    & \textbf{GPQA}    & \textbf{DAPO}
  & \textbf{Avg} \\
\midrule
 
\rowcolor{cGroupBg}[\tabcolsep][\tabcolsep]
\multicolumn{8}{c}{\textbf{Qwen3-4B-Base}} \\[\jot]
Base
    & \scr{10.4}{35.5} & \scr{67.5}{90.2} & \scr{7.8}{31.6}
    & \scr{39.0}{73.3} & \scr{34.2}{83.4} & \scr{28.7}{65.5}
    & \scr{31.3}{63.3} \\
GRPO$^\dagger$
    & \scr{22.0}{45.5} & \scr{81.6}{91.9} & \scr{22.5}{42.5}
    & \scr{57.1}{81.0} & \scr{40.5}{83.4} & \scr{49.9}{77.0}
    & \scr{45.6}{70.2} \\
TTRL
    & \scr{10.8}{14.9} & \scr{72.8}{85.3} & \scr{3.3}{17.6}
    & \scr{42.2}{60.0} & \scr{35.6}{74.9} & \scr{31.2}{55.1}
    & \scr{32.7}{51.3} \\
SCRL
    & \scr{12.1}{25.0} & \scr{75.3}{88.1} & \scr{10.6}{25.4}
    & \scr{46.8}{67.6} & \scr{34.4}{75.8} & \scr{34.7}{60.6}
    & \scr{35.7}{57.1} \\
Co-rewarding
    & \scr{15.9}{26.1} & \scr{77.5}{86.2} & \scr{7.7}{19.9}
    & \scr{49.6}{64.5} & \scr{35.7}{68.7} & \scr{40.1}{60.1}
    & \scr{37.8}{54.3} \\
\rowcolor{cyan!10}[\tabcolsep][\tabcolsep]
\textbf{Ours}
    & \bscr{20.1}{39.0} & \bscr{81.6}{93.2} & \bscr{18.2}{47.0}
    & \bscr{53.7}{80.8} & \bscr{38.1}{84.0} & \bscr{43.0}{76.0}
    & \bscr{42.5}{70.0} \\
\rowcolor{green!10}[\tabcolsep][\tabcolsep]
$\boldsymbol{\Delta}$\,(Ours\,--\,TTRL)
    & \DC{9.3}{24.1}{86}{161}
    & \DC{8.8}{7.9}{12}{9}
    & \DC{14.9}{29.4}{451}{167}
    & \DC{11.5}{20.8}{27}{34}
    & \DC{2.5}{9.1}{7}{12}
    & \DC{11.8}{20.9}{37}{37}
    & \DC{9.8}{18.7}{30}{36} \\
 
\midrule
 
\rowcolor{cGroupBg}[\tabcolsep][\tabcolsep]
\multicolumn{8}{c}{\textbf{Qwen3-8B-Base}} \\[\jot]
Base
    & \scr{11.0}{35.4} & \scr{63.3}{91.2} & \scr{9.1}{31.5}
    & \scr{38.5}{79.6} & \scr{35.3}{88.9} & \scr{29.3}{69.9}
    & \scr{31.1}{66.1} \\
GRPO$^\dagger$
    & \scr{26.5}{55.3} & \scr{85.8}{94.9} & \scr{22.3}{38.8}
    & \scr{65.1}{86.3} & \scr{48.1}{88.3} & \scr{55.7}{81.5}
    & \scr{50.6}{74.2} \\
TTRL
    & \scr{13.6}{28.4} & \scr{77.0}{87.7} & \scr{9.9}{24.3}
    & \scr{50.6}{72.9} & \scr{39.1}{79.6} & \scr{38.7}{64.3}
    & \scr{38.2}{59.5} \\
SCRL
    & \scr{13.5}{32.6} & \scr{77.8}{87.9} & \scr{10.7}{26.4}
    & \scr{51.1}{71.2} & \scr{38.2}{82.5} & \scr{38.4}{63.0}
    & \scr{38.3}{60.6} \\
Co-rewarding
    & \scr{15.9}{30.5} & \scr{80.5}{89.5} & \scr{12.5}{21.0}
    & \scr{54.9}{74.3} & \scr{41.5}{75.6} & \scr{44.7}{63.9}
    & \scr{41.7}{59.1} \\
\rowcolor{cyan!10}[\tabcolsep][\tabcolsep]
\textbf{Ours}
    & \bscr{22.0}{50.5} & \bscr{82.5}{94.6} & \bscr{16.3}{37.9}
    & \bscr{56.4}{84.3} & \bscr{43.6}{89.9} & \bscr{45.3}{77.6}
    & \bscr{44.4}{72.5} \\
\rowcolor{green!10}[\tabcolsep][\tabcolsep]
$\boldsymbol{\Delta}$\,(Ours\,--\,TTRL)
    & \DC{8.4}{22.1}{61}{77}
    & \DC{5.5}{6.9}{7}{7}
    & \DC{6.4}{13.6}{64}{56}
    & \DC{5.8}{11.4}{11}{15}
    & \DC{4.5}{10.3}{11}{12}
    & \DC{6.6}{13.3}{17}{20}
    & \DC{6.2}{13.0}{16}{21} \\
 
\bottomrule
\end{tabular}
}

\end{table}
\subsection{Experimental Results}
\textbf{TTRL-CoCoV significantly outperforms all label-free baselines across pass@1 and pass@16 metrics while mitigating the exploration degradation inherent in standard TTRL.} As shown in Table \ref{tab:main}, standard TTRL often suffers from late-stage mode collapse, causing its pass@16 performance to sharply fall behind the Base model (e.g., dropping from 31.6\% to 17.6\% on AIME25 for Qwen3-4B). In contrast, TTRL-CoCoV sustains diverse reasoning exploration, effectively preventing such performance drops. Averaged across all six benchmarks, our method yields absolute gains of +9.8\% and +18.7\% in pass@1 and pass@16 over TTRL on the Qwen3-4B architecture, alongside robust improvements of +6.2\% and +13.0\% on the larger Qwen3-8B. 

\begin{wrapfigure}[12]{r}[0pt]{0.45\textwidth} 
    \vspace{-12pt}
    \centering
    \includegraphics[width=0.9\linewidth]{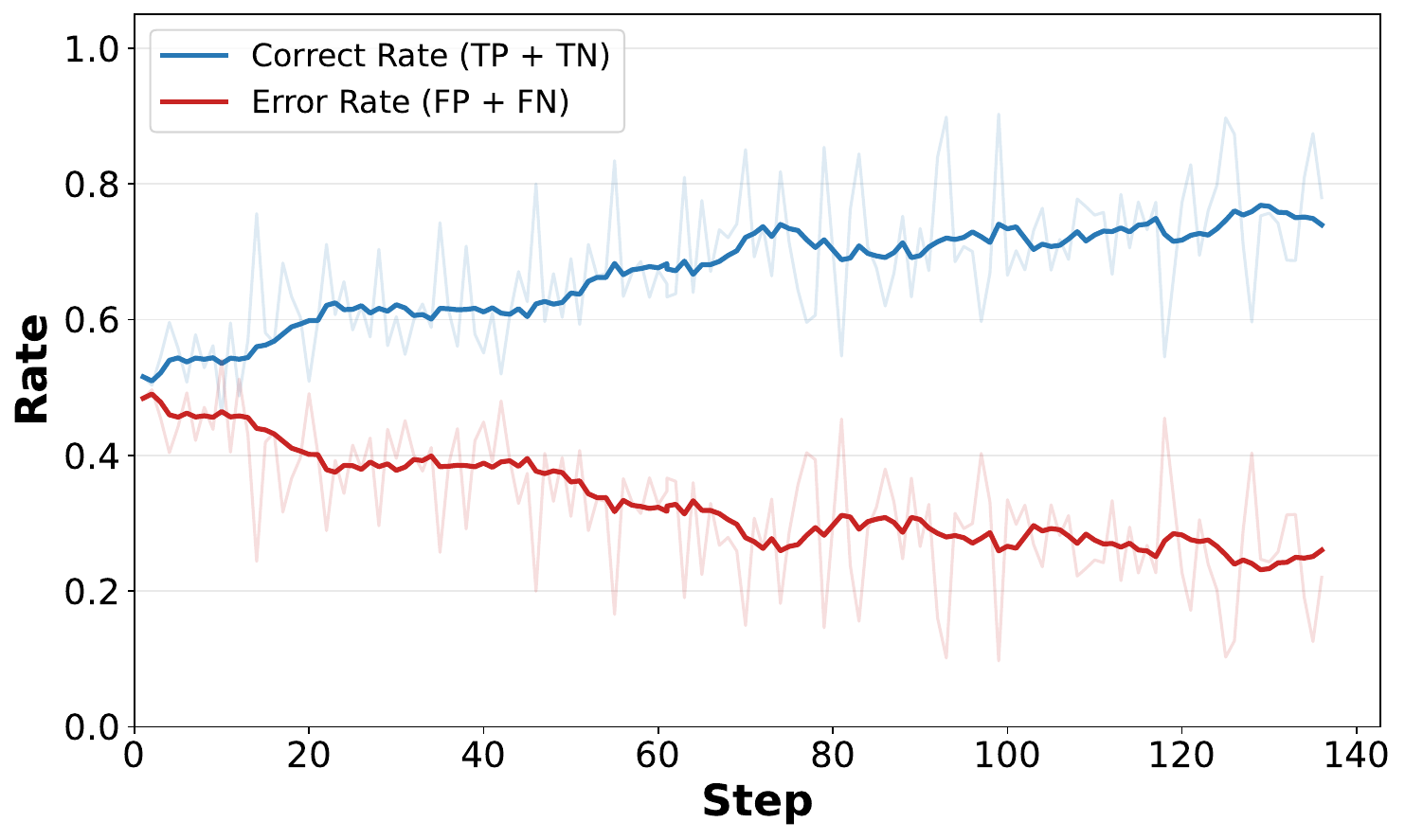}
    \vspace{-6pt} 
    \caption{\small Training dynamics of Verifier: validation correct rate increases while error rate declines.}
    \label{fig:correct_error_rates}
    \vspace{-10pt} 
\end{wrapfigure}

\textbf{TTRL-CoCoV fosters a synergistic enhancement of both generative and verification capabilities.} Tracking the verifier's internal metrics during training (Fig.~\ref{fig:correct_error_rates}) reveals a steady upward trend in verification accuracy, with a consistently converging verification error rate. These dynamics validate the efficacy of our design: the generator produces high-quality reasoning trajectories, while the inherently refined verifier filters out false consensus. Ultimately, this joint parameter update establishes a powerful virtuous cycle of mutual improvement.

\textbf{TTRL-CoCoV approaches or even exceeds the supervised upper bound set by ground-truth labels.} Although TTRL-CoCoV is a purely label-free fine-tuning paradigm, its performance closely approaches the upper bound of fully supervised training with ground-truth labels (GRPO). For instance, on the Qwen3-4B architecture, TTRL-CoCoV achieves an average Pass@16 of 70.0\% across six benchmarks, nearly matching the GRPO upper bound of 70.2\%. More remarkably, it surpasses this supervised ceiling on highly complex tasks, attaining a 47.0\% Pass@16 on AIME25 (vs. GRPO's 42.5\%) and matching GRPO's 81.6\% Pass@1 on MATH500. This substantially bridges the gap between label-free and label-supervised learning.

\begin{wraptable}{r}{0.6\textwidth} 
    \centering
    \small 

    \caption{\small Pass@1 performance comparison with policy annealing. $^*$ denotes our method further optimized with annealing mechanism.}
    \label{tab:annealing}
    \begin{tabular}{lcccc} 
        \toprule
        Method & AIME24 & MATH500 & GPQA & Avg \\
        \midrule
        GRPO & 22.0 & 81.6 & 40.5 & 48.0 \\
        Ours & 20.1 & 81.6 & 38.1 & 46.6 \\
        \rowcolor{cyan!10}[\tabcolsep][\tabcolsep]
        \textbf{Ours}$^*$ & \textbf{23.4} & \textbf{82.6} & \textbf{45.5} & \textbf{50.5} \\
        \midrule
        \rowcolor{green!10}[\tabcolsep][\tabcolsep]
        $\Delta$ (Ours$^*$ - GRPO)& \textcolor{red}{\textbf{+1.4}} & \textcolor{red}{\textbf{+1.0}} & \textcolor{red}{\textbf{+5.0}} & \textcolor{red}{\textbf{+2.5}} \\
        \bottomrule
    \end{tabular}

\end{wraptable}

\textbf{Post-TTRL-CoCoV policy annealing triggers deep exploitation to surpass performance upper bounds.} To investigate whether the model can achieve further convergence after accumulating sufficient exploratory experience, we introduce a two-stage "exploration-to-exploitation" policy annealing mechanism. Through TTRL-CoCoV training, the model constructs a highly diverse pool of correct reasoning trajectories. In the subsequent training phase, we remove the diversity reward and strictly transition the optimization objective toward a single-path Pass@1 greedy maximization (i.e., the exploitation phase). Empirical results demonstrate that this brief convergence phase triggers a secondary surge in model capability  (Table~\ref{tab:annealing}). Not only does it overcome the performance bottleneck of the initial exploratory stage, but it also successfully surpasses the Pass@1 performance of the fully-supervised GRPO across multiple benchmarks. By adopting this "explore-then-exploit" annealing paradigm, our approach effectively bridges the gap between label-free and label-supervised learning, ultimately exceeding the established supervised performance upper bounds.

\begin{figure}[htbp] 
\label{Q123}
  \centering
  \includegraphics[width=0.90\linewidth]{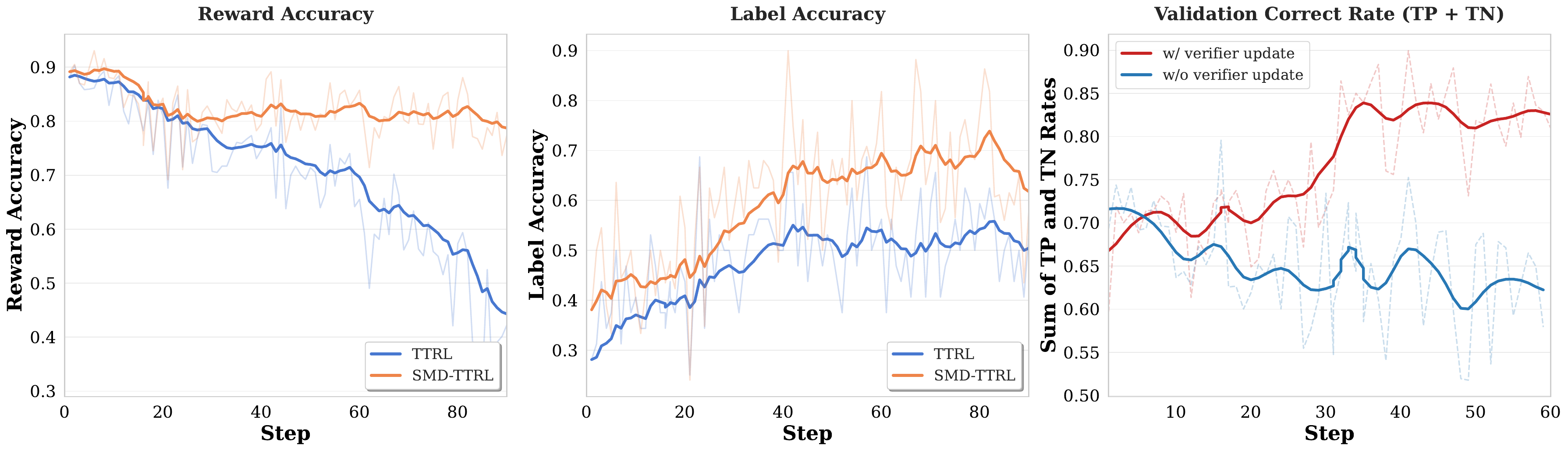}

  \caption{\small Training and internal verification dynamics of TTRL-CoCoV. \textbf{(Left \& Middle):} Reward Accuracy and Label Accuracy. While standard TTRL suffers from late-stage pseudo-label collapse and confirmation bias, TTRL-CoCoV maintains highly stable reward accuracy ($>0.8$) and smoothly increasing label accuracy. \textbf{(Right):} Validation Correction Rate demonstrating co-evolution. Under joint updates, the correct verification rate steadily climbs, whereas a static verifier suffers from capability mismatch and degrades.}
  \label{fig:combined_dynamics}

\end{figure}
\section{Analysis and Discussions}
\label{analysis}

\textbf{Q1: Can TTRL-CoCoV consistently provide high-quality training signals?}
To investigate whether TTRL-CoCoV can overcome the common pseudo-label noise problem in label-free learning, we track the reward accuracy and pseudo-label accuracy throughout the training process (see Fig.~\ref{fig:combined_dynamics}, left and middle). We demonstrate that our confidence-conditioned self-classification and self-verification mechanisms effectively break the vicious cycle of confirmation bias. By accurately filtering out false high-consensus noise, TTRL-CoCoV continuously distills high-quality training signals and completely avoids the late-stage reward collapse seen in traditional TTRL (a detailed numerical analysis is provided in Appendix \ref{sec:appendix_noise}).

\textbf{Q2: How does the co-evolution mechanism in TTRL-CoCoV affect training?}
To investigate the interaction between generation and verification capabilities under shared weights, we established an ablation baseline with a frozen verifier and analyzed the internal verification dynamics (see Fig.~\ref{fig:combined_dynamics}, right). We demonstrate that without co-evolution, a static verifier suffers from a severe capability disconnect as the generator evolves, ultimately becoming a performance bottleneck. Conversely, joint updates create a virtuous cycle: the generator's output continuously calibrates the verifier's discriminative intuition, while the upgraded verifier improves reward accuracy for the generator. This achieves true co-evolution and prevents the significant degradation in downstream exploration capabilities caused by a static verifier (detailed downstream task performance and numerical error metrics are provided in Appendix \ref{sec:appendix_coevolution}).

\textbf{Q3: Why does the length-diversity reward in TTRL-CoCoV enhance the model's exploration capability? }
In traditional TTRL, when the model achieves high consistency on a particular problem, it tends to exploit shortcut learning by outputting short, templated answers to quickly obtain rewards. Our analysis reveals that without the length diversity reward ($R_{div}$), the generated responses become highly homogenized. By introducing $R_{div}$, we successfully force the model out of this comfort zone. Rewarding correct trajectories that deviate from the mean length effectively stimulates the model's exploration capability, enabling it to explore different problem-solving approaches and generate more diverse chains of thought while maintaining high accuracy (detailed length dynamics and variance curves are provided in Appendix \ref{sec:appendix_diversity}).

\textbf{Q4: How do different model sizes and types affect the effectiveness of TTRL-CoCoV?}
To comprehensively evaluate the scalability and generalization of our proposed method, we conduct comparative experiments across varying parameter scales and diverse model architectures. We demonstrate that our framework aligns with reinforcement learning scaling laws, with performance gains increasing significantly as model size grows due to the larger models' stronger inherent reasoning and verification capabilities. Furthermore, cross-model evaluations reveal that while our method achieves stable improvements across different architectures, the magnitude of these gains is highly dependent on domain knowledge: base models with richer prior expertise in the target domain unlock substantially higher exploration and verification gains (comprehensive benchmark evaluations and cross-model comparisons are detailed in Appendices \ref{sec:appendix_scalability} and \ref{sec:appendix_generalization}).

\textbf{Q5: Why TTRL-CoCoV uses an asymmetric soft penalty matrix?}
To validate the design of our asymmetric soft penalty reward matrix during verification, we evaluate its noise-filtering capability against a standard symmetric reward baseline. We demonstrate that while a symmetric setting allows persistently elevated false positive ratios that corrupt gradient signals, our asymmetric strategy—assigning a higher penalty weight to false positives—imposes a rigorous screening criterion. This strictly mitigates the impact of erroneous gradients on the first-stage policy, effectively enhancing overall reward accuracy in label-free learning (detailed verification dynamics and false positive ratio comparisons are provided in Appendix \ref{sec:appendix_asymmetric}).

\section{Conclusion}
\vspace{-0.8em}
We presented TTRL-CoCoV to address the previously under-explored challenge of Pass@k optimization in unsupervised TTRL. By combining an adaptive co-evolutionary verification mechanism with an exploration-enhancing reward, the framework simultaneously resolves pseudo-label unreliability and trajectory diversity collapse. TTRL-CoCoV not only yields consistent gains in generation coverage, but also leverages the better exploratory capability to drive further exploitation, ultimately surpassing the performance ceiling of fully supervised baselines on challenging benchmarks.

\medskip
{
\small
\bibliographystyle{unsrtnat} 
\bibliography{references} 

\begin{thebibliography}{39}
\providecommand{\natexlab}[1]{#1}
\providecommand{\url}[1]{\texttt{#1}}
\expandafter\ifx\csname urlstyle\endcsname\relax
  \providecommand{\doi}[1]{doi: #1}\else
  \providecommand{\doi}{doi: \begingroup \urlstyle{rm}\Url}\fi

\bibitem[Zuo et~al.(2025)Zuo, Zhang, Sheng, Qu, Cui, Zhu, Li, Zhang, Long, Hua, et~al.]{zuo2025ttrl}
Yuxin Zuo, Kaiyan Zhang, Li~Sheng, Shang Qu, Ganqu Cui, Xuekai Zhu, Haozhan Li, Yuchen Zhang, Xinwei Long, Ermo Hua, et~al.
\newblock Ttrl: Test-time reinforcement learning.
\newblock \emph{arXiv preprint arXiv:2504.16084}, 2025.

\bibitem[Liu et~al.(2025)Liu, He, Lin, Yang, Shen, and Liu]{liu2025ettrl}
Jia Liu, ChangYi He, YingQiao Lin, MingMin Yang, FeiYang Shen, and ShaoGuo Liu.
\newblock Ettrl: Balancing exploration and exploitation in llm test-time reinforcement learning via entropy mechanism.
\newblock \emph{arXiv preprint arXiv:2508.11356}, 2025.

\bibitem[Wang et~al.(2025{\natexlab{a}})Wang, Huang, Cao, Iwasawa, Matsuo, and Guo]{wang2025self}
Ru~Wang, Wei Huang, Qi~Cao, Yusuke Iwasawa, Yutaka Matsuo, and Jiaxian Guo.
\newblock Self-harmony: Learning to harmonize self-supervision and self-play in test-time reinforcement learning.
\newblock \emph{arXiv preprint arXiv:2511.01191}, 2025{\natexlab{a}}.

\bibitem[Zhou et~al.(2025)Zhou, Liang, Liu, Yu, Panaganti, Song, Yu, Zhang, Mi, and Yu]{zhou2025evolving}
Yujun Zhou, Zhenwen Liang, Haolin Liu, Wenhao Yu, Kishan Panaganti, Linfeng Song, Dian Yu, Xiangliang Zhang, Haitao Mi, and Dong Yu.
\newblock Evolving language models without labels: Majority drives selection, novelty promotes variation.
\newblock \emph{arXiv preprint arXiv:2509.15194}, 2025.

\bibitem[Jaech et~al.(2024)Jaech, Kalai, Lerer, Richardson, El-Kishky, Low, Helyar, Madry, Beutel, Carney, et~al.]{jaech2024openai}
Aaron Jaech, Adam Kalai, Adam Lerer, Adam Richardson, Ahmed El-Kishky, Aiden Low, Alec Helyar, Aleksander Madry, Alex Beutel, Alex Carney, et~al.
\newblock Openai o1 system card.
\newblock \emph{arXiv preprint arXiv:2412.16720}, 2024.

\bibitem[Guo et~al.(2025)Guo, Yang, Zhang, Song, Wang, Zhu, Xu, Zhang, Ma, Bi, et~al.]{guo2025deepseek}
Daya Guo, Dejian Yang, Haowei Zhang, Junxiao Song, Peiyi Wang, Qihao Zhu, Runxin Xu, Ruoyu Zhang, Shirong Ma, Xiao Bi, et~al.
\newblock Deepseek-r1: Incentivizing reasoning capability in llms via reinforcement learning.
\newblock \emph{arXiv preprint arXiv:2501.12948}, 2025.

\bibitem[Shao et~al.(2024)Shao, Wang, Zhu, Xu, Song, Bi, Zhang, Zhang, Li, Wu, et~al.]{shao2024deepseekmath}
Zhihong Shao, Peiyi Wang, Qihao Zhu, Runxin Xu, Junxiao Song, Xiao Bi, Haowei Zhang, Mingchuan Zhang, YK~Li, Yang Wu, et~al.
\newblock Deepseekmath: Pushing the limits of mathematical reasoning in open language models.
\newblock \emph{arXiv preprint arXiv:2402.03300}, 2024.

\bibitem[Chen et~al.(2025{\natexlab{a}})Chen, Qin, Wu, Ling, Ye, Zhao, and Shi]{chen2025pass}
Zhipeng Chen, Xiaobo Qin, Youbin Wu, Yue Ling, Qinghao Ye, Wayne~Xin Zhao, and Guang Shi.
\newblock Pass@ k training for adaptively balancing exploration and exploitation of large reasoning models.
\newblock \emph{arXiv preprint arXiv:2508.10751}, 2025{\natexlab{a}}.

\bibitem[Walder and Karkhanis(2025)]{walder2025pass}
Christian Walder and Deep Karkhanis.
\newblock Pass@ k policy optimization: Solving harder reinforcement learning problems.
\newblock \emph{arXiv preprint arXiv:2505.15201}, 2025.

\bibitem[Yan et~al.(2026)Yan, Liang, Wang, Lu, He, and Tan]{yan2026if}
Dong Yan, Jian Liang, Yanbo Wang, Shuo Lu, Ran He, and Tieniu Tan.
\newblock What if consensus lies? selective-complementary reinforcement learning at test time.
\newblock \emph{arXiv preprint arXiv:2603.19880}, 2026.

\bibitem[Yu et~al.(2025{\natexlab{a}})Yu, Su, Tao, Wang, Singh, Yu, Wang, Gao, Yuan, Weston, et~al.]{yu2025restrain}
Zhaoning Yu, Will Su, Leitian Tao, Haozhu Wang, Aashu Singh, Hanchao Yu, Jianyu Wang, Hongyang Gao, Weizhe Yuan, Jason Weston, et~al.
\newblock Restrain: From spurious votes to signals--self-driven rl with self-penalization.
\newblock \emph{arXiv preprint arXiv:2510.02172}, 2025{\natexlab{a}}.

\bibitem[Pan et~al.(2026)Pan, Yan, Wang, Zhang, Han, Zhang, Lu, Xiao, and Shen]{pan2026coverrl}
Teng Pan, Yuchen Yan, Zixuan Wang, Ruiqing Zhang, Gaiyang Han, Wanqi Zhang, Weiming Lu, Jun Xiao, and Yongliang Shen.
\newblock Coverrl: Breaking the consensus trap in label-free reasoning via generator-verifier co-evolution.
\newblock \emph{arXiv preprint arXiv:2603.17775}, 2026.

\bibitem[Liao et~al.(2026)Liao, R{\"o}hrich, Wang, Zhang, Samadzadeh, Tresp, and Yeung-Levy]{liao2026tool}
Ruotong Liao, Nikolai R{\"o}hrich, Xiaohan Wang, Yuhui Zhang, Yasaman Samadzadeh, Volker Tresp, and Serena Yeung-Levy.
\newblock Tool verification for test-time reinforcement learning.
\newblock \emph{arXiv preprint arXiv:2603.02203}, 2026.

\bibitem[Wan et~al.(2026)Wan, Shen, Dou, Zhou, Zhang, Wang, Shen, Xiong, Tao, Zhong, et~al.]{wan2026dsdr}
Zhongwei Wan, Yun Shen, Zhihao Dou, Donghao Zhou, Yu~Zhang, Xin Wang, Hui Shen, Jing Xiong, Chaofan Tao, Zixuan Zhong, et~al.
\newblock Dsdr: Dual-scale diversity regularization for exploration in llm reasoning.
\newblock \emph{arXiv preprint arXiv:2602.19895}, 2026.

\bibitem[Zhang et~al.(2025{\natexlab{a}})Zhang, Zhu, Ge, Zhao, Zhou, Li, Feng, Yao, and Han]{zhang2025co}
Zizhuo Zhang, Jianing Zhu, Xinmu Ge, Zihua Zhao, Zhanke Zhou, Xuan Li, Xiao Feng, Jiangchao Yao, and Bo~Han.
\newblock Co-rewarding: Stable self-supervised rl for eliciting reasoning in large language models.
\newblock \emph{arXiv preprint arXiv:2508.00410}, 2025{\natexlab{a}}.

\bibitem[Zelikman et~al.(2022)Zelikman, Wu, Mu, and Goodman]{zelikman2022star}
Eric Zelikman, Yuhuai Wu, Jesse Mu, and Noah Goodman.
\newblock Star: Bootstrapping reasoning with reasoning.
\newblock \emph{Advances in Neural Information Processing Systems}, 35:\penalty0 15476--15488, 2022.

\bibitem[Yuan et~al.(2024)Yuan, Pang, Cho, Li, Sukhbaatar, Xu, and Weston]{yuan2024self}
Weizhe Yuan, Richard~Yuanzhe Pang, Kyunghyun Cho, Xian Li, Sainbayar Sukhbaatar, Jing Xu, and Jason Weston.
\newblock Self-rewarding language models.
\newblock \emph{arXiv preprint arXiv:2401.10020}, 2024.

\bibitem[Shafayat et~al.(2025)Shafayat, Tajwar, Salakhutdinov, Schneider, and Zanette]{shafayat2025can}
Sheikh Shafayat, Fahim Tajwar, Ruslan Salakhutdinov, Jeff Schneider, and Andrea Zanette.
\newblock Can large reasoning models self-train?
\newblock \emph{arXiv preprint arXiv:2505.21444}, 2025.

\bibitem[Du et~al.(2026)Du, Huang, and Li]{du2026distribution}
Bodong Du, Xuanqi Huang, and Xiaomeng Li.
\newblock Distribution-aware reward estimation for test-time reinforcement learning.
\newblock \emph{arXiv preprint arXiv:2601.21804}, 2026.

\bibitem[Lightman et~al.(2023)Lightman, Kosaraju, Burda, Edwards, Baker, Lee, Leike, Schulman, Sutskever, and Cobbe]{lightman2023let}
Hunter Lightman, Vineet Kosaraju, Yuri Burda, Harrison Edwards, Bowen Baker, Teddy Lee, Jan Leike, John Schulman, Ilya Sutskever, and Karl Cobbe.
\newblock Let's verify step by step.
\newblock In \emph{The twelfth international conference on learning representations}, 2023.

\bibitem[Weng et~al.(2023)Weng, Zhu, Xia, Li, He, Liu, Sun, Liu, and Zhao]{weng2023large}
Yixuan Weng, Minjun Zhu, Fei Xia, Bin Li, Shizhu He, Shengping Liu, Bin Sun, Kang Liu, and Jun Zhao.
\newblock Large language models are better reasoners with self-verification.
\newblock In \emph{Findings of the Association for Computational Linguistics: EMNLP 2023}, pages 2550--2575, 2023.

\bibitem[Zhao et~al.(2025)Zhao, Wu, Yue, Wu, Xu, Lin, Wang, Wu, Zheng, and Huang]{zhao2025absolute}
Andrew Zhao, Yiran Wu, Yang Yue, Tong Wu, Quentin Xu, Matthieu Lin, Shenzhi Wang, Qingyun Wu, Zilong Zheng, and Gao Huang.
\newblock Absolute zero: Reinforced self-play reasoning with zero data.
\newblock \emph{arXiv preprint arXiv:2505.03335}, 2025.

\bibitem[Huang et~al.(2025)Huang, Yu, Wang, Zhang, Li, Li, Huang, Mi, and Yu]{huang2025r}
Chengsong Huang, Wenhao Yu, Xiaoyang Wang, Hongming Zhang, Zongxia Li, Ruosen Li, Jiaxin Huang, Haitao Mi, and Dong Yu.
\newblock R-zero: Self-evolving reasoning llm from zero data.
\newblock \emph{arXiv preprint arXiv:2508.05004}, 2025.

\bibitem[Chen et~al.(2025{\natexlab{b}})Chen, Zhang, Ma, Wang, Liang, Tu, Li, and Wong]{chen2025spc}
Jiaqi Chen, Bang Zhang, Ruotian Ma, Peisong Wang, Xiaodan Liang, Zhaopeng Tu, Xiaolong Li, and Kwan-Yee~K Wong.
\newblock Spc: Evolving self-play critic via adversarial games for llm reasoning.
\newblock \emph{arXiv preprint arXiv:2504.19162}, 2025{\natexlab{b}}.

\bibitem[Zhang et~al.(2025{\natexlab{b}})Zhang, Huang, Li, and Cardie]{zhang2025better}
Zhengxin Zhang, Chengyu Huang, Aochong~Oliver Li, and Claire Cardie.
\newblock Better llm reasoning via dual-play.
\newblock \emph{arXiv preprint arXiv:2511.11881}, 2025{\natexlab{b}}.

\bibitem[Zhang et~al.(2025{\natexlab{c}})Zhang, Xu, Wang, Cui, Liu, and An]{zhang2025incentivizing}
Fuxiang Zhang, Jiacheng Xu, Chaojie Wang, Ce~Cui, Yang Liu, and Bo~An.
\newblock Incentivizing llms to self-verify their answers.
\newblock \emph{arXiv preprint arXiv:2506.01369}, 2025{\natexlab{c}}.

\bibitem[Chen et~al.(2026)Chen, Wang, Zhang, Ye, Cai, Shi, Gu, Su, Cai, Wang, et~al.]{chen2026learning}
Yuxin Chen, Yu~Wang, Yi~Zhang, Ziang Ye, Zhengzhou Cai, Yaorui Shi, Qi~Gu, Hui Su, Xunliang Cai, Xiang Wang, et~al.
\newblock Learning to self-verify makes language models better reasoners.
\newblock \emph{arXiv preprint arXiv:2602.07594}, 2026.

\bibitem[Song et~al.(2025)Song, Kempe, and Munos]{song2025outcome}
Yuda Song, Julia Kempe, and Remi Munos.
\newblock Outcome-based exploration for llm reasoning.
\newblock \emph{arXiv preprint arXiv:2509.06941}, 2025.

\bibitem[Bi et~al.(2024)Bi, Han, Liu, Tang, and Wang]{bi2024forest}
Zhenni Bi, Kai Han, Chuanjian Liu, Yehui Tang, and Yunhe Wang.
\newblock Forest-of-thought: Scaling test-time compute for enhancing llm reasoning.
\newblock \emph{arXiv preprint arXiv:2412.09078}, 2024.

\bibitem[Khanh et~al.(2026)Khanh, Zhu, Yue, and Nguyen]{khanh2026test}
Ly~Tran~Ho Khanh, Dongxuan Zhu, Man-Chung Yue, and Viet~Anh Nguyen.
\newblock Test-time diverse reasoning by riemannian activation steering.
\newblock In \emph{Proceedings of the AAAI Conference on Artificial Intelligence}, volume~40, pages 31429--31437, 2026.

\bibitem[Wu et~al.(2025)Wu, George, Ye, Wu, Schmidt, and Cai]{wu2025spine}
Jianghao Wu, Yasmeen George, Jin Ye, Yicheng Wu, Daniel~F Schmidt, and Jianfei Cai.
\newblock Spine: Token-selective test-time reinforcement learning with entropy-band regularization.
\newblock \emph{arXiv preprint arXiv:2511.17938}, 2025.

\bibitem[Yang et~al.(2025)Yang, Li, Yang, Zhang, Hui, Zheng, Yu, Gao, Huang, Lv, et~al.]{yang2025qwen3}
An~Yang, Anfeng Li, Baosong Yang, Beichen Zhang, Binyuan Hui, Bo~Zheng, Bowen Yu, Chang Gao, Chengen Huang, Chenxu Lv, et~al.
\newblock Qwen3 technical report.
\newblock \emph{arXiv preprint arXiv:2505.09388}, 2025.

\bibitem[Yang et~al.(2024)Yang, Zhang, Hui, Gao, Yu, Li, Liu, Tu, Zhou, Lin, et~al.]{yang2024qwen2}
An~Yang, Beichen Zhang, Binyuan Hui, Bofei Gao, Bowen Yu, Chengpeng Li, Dayiheng Liu, Jianhong Tu, Jingren Zhou, Junyang Lin, et~al.
\newblock Qwen2. 5-math technical report: Toward mathematical expert model via self-improvement.
\newblock \emph{arXiv preprint arXiv:2409.12122}, 2024.

\bibitem[Wang et~al.(2025{\natexlab{b}})Wang, Zhou, Li, and Liu]{wang2025octothinker}
Zengzhi Wang, Fan Zhou, Xuefeng Li, and Pengfei Liu.
\newblock Octothinker: Mid-training incentivizes reinforcement learning scaling.
\newblock \emph{arXiv preprint arXiv:2506.20512}, 2025{\natexlab{b}}.

\bibitem[Grattafiori et~al.(2024)Grattafiori, Dubey, Jauhri, Pandey, Kadian, Al-Dahle, Letman, Mathur, Schelten, Vaughan, et~al.]{grattafiori2024llama}
Aaron Grattafiori, Abhimanyu Dubey, Abhinav Jauhri, Abhinav Pandey, Abhishek Kadian, Ahmad Al-Dahle, Aiesha Letman, Akhil Mathur, Alan Schelten, Alex Vaughan, et~al.
\newblock The llama 3 herd of models.
\newblock \emph{arXiv preprint arXiv:2407.21783}, 2024.

\bibitem[Yu et~al.(2025{\natexlab{b}})Yu, Zhang, Zhu, Yuan, Zuo, Yue, Dai, Fan, Liu, Liu, et~al.]{yu2025dapo}
Qiying Yu, Zheng Zhang, Ruofei Zhu, Yufeng Yuan, Xiaochen Zuo, Yu~Yue, Weinan Dai, Tiantian Fan, Gaohong Liu, Lingjun Liu, et~al.
\newblock Dapo: An open-source llm reinforcement learning system at scale, 2025.
\newblock \emph{URL https://arxiv. org/abs/2503.14476}, 1:\penalty0 2, 2025{\natexlab{b}}.

\bibitem[Hendrycks et~al.(2021)Hendrycks, Burns, Kadavath, Arora, Basart, Tang, Song, and Steinhardt]{hendrycks2021measuring}
Dan Hendrycks, Collin Burns, Saurav Kadavath, Akul Arora, Steven Basart, Eric Tang, Dawn Song, and Jacob Steinhardt.
\newblock Measuring mathematical problem solving with the math dataset.
\newblock \emph{arXiv preprint arXiv:2103.03874}, 2021.

\bibitem[Li et~al.(2024)Li, Beeching, Tunstall, Lipkin, Soletskyi, Huang, Rasul, Yu, Jiang, Shen, et~al.]{li2024numinamath}
Jia Li, Edward Beeching, Lewis Tunstall, Ben Lipkin, Roman Soletskyi, Shengyi Huang, Kashif Rasul, Longhui Yu, Albert~Q Jiang, Ziju Shen, et~al.
\newblock Numinamath: The largest public dataset in ai4maths with 860k pairs of competition math problems and solutions.
\newblock \emph{Hugging Face repository}, 13\penalty0 (9):\penalty0 9, 2024.

\bibitem[Rein et~al.(2023)Rein, Hou, Stickland, Petty, Pang, Dirani, Michael, and Bowman]{rein2023gpqa}
David Rein, Betty~Li Hou, Asa~Cooper Stickland, Jackson Petty, Richard~Yuanzhe Pang, Julien Dirani, Julian Michael, and Samuel~R Bowman.
\newblock Gpqa: A graduate-level google-proof q\&a benchmark.
\newblock \emph{arXiv preprint arXiv:2311.12022}, 2023.

\end{thebibliography}
}

\clearpage
\appendix

\section*{Appendix}
\section{Related Work}
\textbf{Label-free test-time reinforcement learning.} STaR \citep{zelikman2022star} and SRLMs \citep{yuan2024self} established the foundation of annotation-free self-improvement. Building on this, TTRL \citep{zuo2025ttrl} formalized majority-vote consensus over self-sampled rollouts as a general unsupervised fine-tuning paradigm. Subsequent works further strengthened this framework along the dimension of label reliability: SRT \citep{shafayat2025can} via consistency-based filtering, DARE \citep{du2026distribution} via distributional reward estimation, RESTRAIN \citep{yu2025restrain} via self-penalization, SCRL \citep{yan2026if} via complementary positive-negative labeling, and Co-rewarding \citep{zhang2025co} via cross-view supervision. Our work complements this line by introducing confidence stratification to jointly address pseudo-label quality and trajectory diversity within a unified framework.

\textbf{Synergistic generation and verification.} A related line of research uses the asymmetry that verification is easier than generation to construct reliable training signals without external annotation. Early verifier-based methods (\citep{lightman2023let},\citep{weng2023large}) relied on dedicated reward models, while more recent work unifies generation and verification within a single model through self-play, adversarial games, or cooperative RL objectives (\citep{zhao2025absolute},\citep{huang2025r},\citep{chen2025spc},\citep{zhang2025better}). Self-Harmony \citep{wang2025self}, Incentivizing LLMs to Self-Verify \citep{zhang2025incentivizing}, and Learning to Self-Verify \citep{chen2026learning} further investigate conditions under which self-verification yields reliable signals, focusing on training stability, distribution alignment, and over-verification suppression, respectively. CoVerRL \citep{pan2026coverrl} represents a particularly relevant instantiation, bootstrapping both roles through shared weights in a co-evolutionary TTRL loop. Our work extends this direction by introducing confidence-conditioned verification allocation.

\textbf{Exploration and diversity in LLM reasoning.} In standard training-time RL, methods such as Outcome-based Exploration \citep{song2025outcome}, PKPO \citep{walder2025pass}, Pass@k Training \citep{chen2025pass}, and DSDR \citep{wan2026dsdr} promote diverse solution strategies with access to ground-truth labels. At pure inference time, FoT \citep{bi2024forest} and SPREAD \citep{khanh2026test} improve output diversity without parameter updates. In the label-free test-time RL setting, SPINE \citep{wu2025spine} and Evol-RL \citep{zhou2025evolving} explore diversity-aware objectives, representing early steps toward sustained exploration under unsupervised conditions. Our work builds on this direction by explicitly studying diversity collapse as a distinct failure mode and proposing a confidence-conditioned remedy.

\clearpage
\section{Algorithmic Pseudocode of TTRL-CoCoV}
This section presents the algorithmic pseudocode for the TTRL-CoCoV training loop, detailing the conditional verification process and trajectory generation based on majority confidence.

\begin{algorithm}[tbh]
\caption{TTRL-CoCoV Training Loop}
\label{alg:smd_ttrl}
\begin{algorithmic}[1]
\Require Pretrained model $\pi_\theta$, dataset $\mathcal{D}$, thresholds $\tau_{high}, \tau_{low}$, candidate size $K$, verification attempts $m$
\For{each batch $(x)$ in $\mathcal{D}$}
    \State Generate $N$ reasoning trajectories $\{y_1, \dots, y_N\} \sim \pi_\theta(\cdot|x)$
    \State Compute majority confidence $C_{maj}$ and extract Top-$K$ candidates $\{o_k\}_{k=1}^{K}$
    
    \If{$C_{maj} \ge \tau_{high}$} \Comment{High-Consistency}
        \State $\mathbb{I}_{verify} \gets 1; y^* \gets y_{maj}$
        \State Calculate $\mathcal{L}_{first}$ with Length-Diversity Reward $R_{div}$
    \ElsIf{$\tau_{low} \le C_{maj} < \tau_{high}$} \Comment{Mid-Consistency}
        \State $\mathbb{I}_{verify} \gets 0; y^* \gets y_{maj}$
        \State Calculate baseline $\mathcal{L}_{first}$ (No verification)
    \Else \Comment{Low-Consistency}
        \State $\mathbb{I}_{verify} \gets 1$; Compute $\mathrm{VPR}(o_k)$ for Top-$K$ via verifier
        \If{$\exists o_k$ with $\mathrm{VPR} > 0.5$}
            \State $y^* \gets$ best candidate in $\text{true\_set}$
        \Else
            \State $\mathcal{L}_{total} \gets 0$; \textbf{Continue} \Comment{Skip update}
        \EndIf
    \EndIf
    
    \If{$\mathbb{I}_{verify} == 1$}
        \State Generate $m$ verification trajectories per candidate
        \State Compute $\mathcal{L}_{second}$ using Asymmetric Rewards matrix
    \EndIf
    
    \State $\mathcal{L}_{total} = \mathcal{L}_{first} + \mathbb{I}_{verify} \cdot \mathcal{C}_{pseudo} \cdot \mathrm{Mean}(\mathcal{L}_{second})$
    \State Update parameters $\theta$ using gradients $\nabla_\theta \mathcal{L}_{total}$
\EndFor
\end{algorithmic}
\end{algorithm}

\clearpage
\section{Implementation Details}
\label{details}

\subsection{RL Hyperparameters}
The specific hyperparameters utilized during the reinforcement learning phase of TTRL-CoCoV are outlined in Table \ref{tab:smd-ttrl-training-settings}. This includes generator, verifier, and PPO-specific configurations.

\begin{table}[htbp]
  \caption{TTRL-CoCoV Training Settings}
  \label{tab:smd-ttrl-training-settings}
  \centering
  \small
  \begin{tabular}{ll}
    \toprule
    Method & Hyperparameters \\
    \midrule
    Generator & $n_{\text{vote}} = 64$ \\
    & $n_{\text{samples\_per\_prompt}} = 32$ \\
    & Top-$p$ = 1.0 \\
    & Training Temperature = 1.0 \\ 
    & $K_{pass}=4$ \\
    \midrule 
    Verifier & Temperature: $T_{high} = 1.0$, $T_{low} = 0.6$ \\
    & $\tau_{high} = 0.6$, $\tau_{low} = 0.4$ \\
    & Top-$K$ candidates: $K_{high} = 3$, $K_{low} = 5$ \\
    & Top-$p$ = 0.85 \\
    & $n_{\text{verification\_samples}} = 8$ \\
    \midrule
    Length Diversity & $\lambda_{\text{div}} = 0.05$, $C_{max} = 2$ \\
    & $\tau_{high} = 0.6$ \\
    \midrule 
    PPO Trainer & Learning Rate = $5 \times 10^{-7}$ \\
    & $\gamma = 1.0$, $\lambda = 1.0$ \\
    & Use KL Loss = True \\
    & KL Coefficient = $0.001$ \\
    \midrule
    Batch Sizes & $\text{train\_batch\_size} = 32$ \\
    & $\text{rollout\_batch\_size} = 32$ \\
    & $\text{mini\_batch\_size} = 1$ \\
    & $\text{micro\_batch\_size} = 8$ \\
    \midrule
    Lengths & Prompt Max Length = 1024 \\
    & Generate Max Length = 4096 \\
    & Verify Max Length = 2048 \\
    \midrule
    Training Schedule & Epochs = 1 \\
    \midrule
    Evaluation & $n_{\text{samples}} = 32$ \\
    & Temperature $T = 0.6$ \\
    & Top-$p = 0.95$, Top-$K = 20$ \\
    & Metrics: Pass@1, Pass@16 \\
    \bottomrule
  \end{tabular}
\end{table}
\subsection{Hardware Setup}
\label{hardware}
All experiments, including the reinforcement learning training phase and subsequent evaluations of the TTRL-CoCoV framework, were conducted on a compute node equipped with 4 $\times$ NVIDIA H100 GPUs.

\definecolor{MyPurple}{RGB}{128, 0, 128}
\subsection{Self Verification Prompt}
\begin{tcolorbox}[
    title=Box B.2: Self Verification Prompt for Mathematical Hypothesis Testing,
    colback=cyan!5!white,
    colframe=cyan!75!black,
    colbacktitle=cyan!75!black,
    coltitle=white,
    fonttitle=\bfseries,
    boxrule=1pt,
    arc=3pt,
    left=10pt, right=10pt, top=10pt, bottom=10pt,
]
Problem:

\textcolor{red}{\{problem\}}

\vspace{1em}
\textbf{[Hypothesis to Test]}

A previous attempt at this problem resulted in the following answer:

\textcolor{red}{\{candidate\_answer\}}

\vspace{1em}
\textbf{[Task]}

Act as a rigorous mathematical reviewer. 
Treat the previous answer (\textcolor{red}{\{candidate\_answer\}}) as a given hypothesis. Plug this answer BACK into the original problem conditions. Perform a rigorous backward-substitution to check if it satisfies all constraints or if it leads to a mathematical contradiction. 

You MUST conclude your verification by explicitly stating either "Verification Result: True" (if the hypothesis perfectly satisfies all conditions) or "Verification Result: False" (if it leads to any contradiction).

You MUST strictly use the following XML format for your response:

\vspace{0.5em}
\texttt{<reverse\_verification>}

(Your step-by-step backward substitution checking if \textcolor{red}{\{candidate\_answer\}} contradicts the problem conditions)

\textcolor{red}{Verification Result: [True/False]}

\texttt{</reverse\_verification>}
\end{tcolorbox}

\section{Additional Experimental Setup}
\label{setup}
This section provides comprehensive details regarding our experimental configurations to ensure full reproducibility. Specifically, we outline the diverse suite of foundational models, the training and evaluation datasets, and the exact sampling procedures used for our metrics.
\paragraph{Models}
For our main experiments, we evaluate the effectiveness of our method using Qwen3-4B-Base and Qwen3-8B-Base \citep{yang2025qwen3} as the foundational models. To further assess scalability and generalizability across varying model capacities and architectures, we conduct auxiliary experiments and ablation studies on a diverse suite of models. These include lightweight variants (Qwen3-0.6B and 1.7B Base models \citep{yang2025qwen3}), a domain-specific model (Qwen2.5-Math-7B \citep{yang2024qwen2}), and OctoThinker-8B-Hybrid-Base \citep{wang2025octothinker}, which is continually pre-trained from Llama3.1-8B \citep{grattafiori2024llama}.

\paragraph{Datasets}
We train our model using DAPO-14k-MATH, an English subset derived from DAPO-Math-17k \citep{yu2025dapo}, comprising 14,000 deduplicated and standardized math reasoning samples with varying difficulty levels. To evaluate performance, we conduct experiments on six widely-recognized benchmarks, including four specialized mathematical reasoning tasks: MATH-500 \citep{hendrycks2021measuring}, AIME24, AIME25, and AMC \citep{li2024numinamath}, as well as two benchmarks for broader capabilities: GPQA-Diamond \citep{rein2023gpqa} and the DAPO-14k-MATH-test set \citep{yu2025dapo}.

\paragraph{Metrics}
For each problem, we independently sample 32 reasoning trajectories with a temperature of 0.6 and a top-$p$ of 0.95, and we compute the following metrics\citep{zhou2025evolving}:
\begin{itemize}
    \item \textbf{pass@1}: The average accuracy across the 32 sampled trajectories.
    \item \textbf{pass@16}: The probability of obtaining at least one correct answer when randomly sampling 16 trajectories (with replacement) from the total of 32, averaged over 1,000 bootstrap iterations.
\end{itemize}
\section{Additional Experimental Results}

\begin{wraptable}{r}{0.6\textwidth} 
\centering
\small 
\caption{Pass@1 vs. Pass@k performance. Standard TTRL severely collapses response diversity (Pass@k), and naively adapting the Pass@k objective (Pass@k TTA) fails to fully recover it due to unmitigated pseudo-label noise.}
\label{tab:insight}
\begin{tabular}{lccc} 
\toprule
\textbf{Method}   & \textbf{AIME24} & \textbf{AIME25} & \textbf{AMC}  \\
\midrule
Qwen3-4B-Base     & 10.4 / 35.5     & 7.8 / 31.6      & 39.0 / 73.3   \\
TTRL              & 10.8 / 14.9     & 3.3 / 17.6      & 42.2 / 60.0   \\
Pass@k TTA        & 11.6 / 27.0     & 8.5 / 24.6      & 47.1 / 72.1   \\
\bottomrule
\end{tabular}
\end{wraptable}

\subsection{Detailed Empirical Analysis of Naive Pass@k Adaptation}
\label{insigt1}
To empirically demonstrate the failure of naively adapting the Pass@k objective in label-free settings, we evaluated a straightforward adaptation (denoted as Pass@k TTA), which treats non-majority candidate responses as $N_{\text{neg}}$, using the Qwen3-4B-Base model across the AIME24, AIME25, and AMC benchmarks. Main results are present in Table \ref{tab:insight}. Our observations reveal that while standard TTRL generally improves single-pass accuracy (Pass@1), it triggers a severe collapse in generation diversity (Pass@k). For instance, on the AIME24 benchmark, the Pass@k metric drops drastically from the base model's 35.5 to 14.9 after standard TTRL training. Furthermore, implementing Pass@k TTA only partially mitigates this degradation—recovering the Pass@k to 27.0 on AIME24—which still falls significantly short of the base model's original exploratory capacity. Similar performance trends were consistently observed on the AIME25 and AMC datasets. These results empirically confirm our core conclusion: without a ground-truth oracle to guarantee label correctness, naively rewarding low-consistency exploratory signals is fundamentally compromised by severe pseudo-label noise, ultimately failing to sustain the required exploratory space for complex reasoning tasks.

\subsection{Detailed Numerical Analysis of Noise Resilience}
\label{sec:appendix_noise}

To further elaborate on the training dynamics presented in Fig.~\ref{fig:combined_dynamics} (left and middle) of the main text, we provide a detailed numerical breakdown of the accuracy trends.

In traditional TTRL, the model is prone to reward collapse in the mid-to-late training stages. The underlying reason is often the overconfidence of the policy: as optimization progresses, the model tends to prematurely converge to erroneous consensus on difficult samples, distorting low-consistency guesses into false high-consensus errors, which causes the reward accuracy to drop sharply to around 0.4.

In contrast, our empirical results show that even in the later stages where baseline methods completely collapse, TTRL-CoCoV maintains its reward accuracy robustly at a high level of 0.8. In terms of label accuracy, our method establishes an early lead of approximately 0.1. As the two roles of generator and verifier co-evolve, this gap further expands to 0.2 in the later stages. Ultimately, our label accuracy converges smoothly to a high level around 0.7. These detailed results demonstrate that TTRL-CoCoV can accurately identify and filter out false high-consensus noise throughout the entire cycle.

\subsection{Detailed Impact of Generator-Verifier Co-evolution}
\label{sec:appendix_coevolution}

To provide further empirical evidence for the necessity of joint updates discussed in the main text, we detail the downstream task performance and specific numerical changes in verification error metrics.

\begin{figure}[h]
    \centering
    \hspace{2cm} 
    \includegraphics[width=0.8\linewidth]{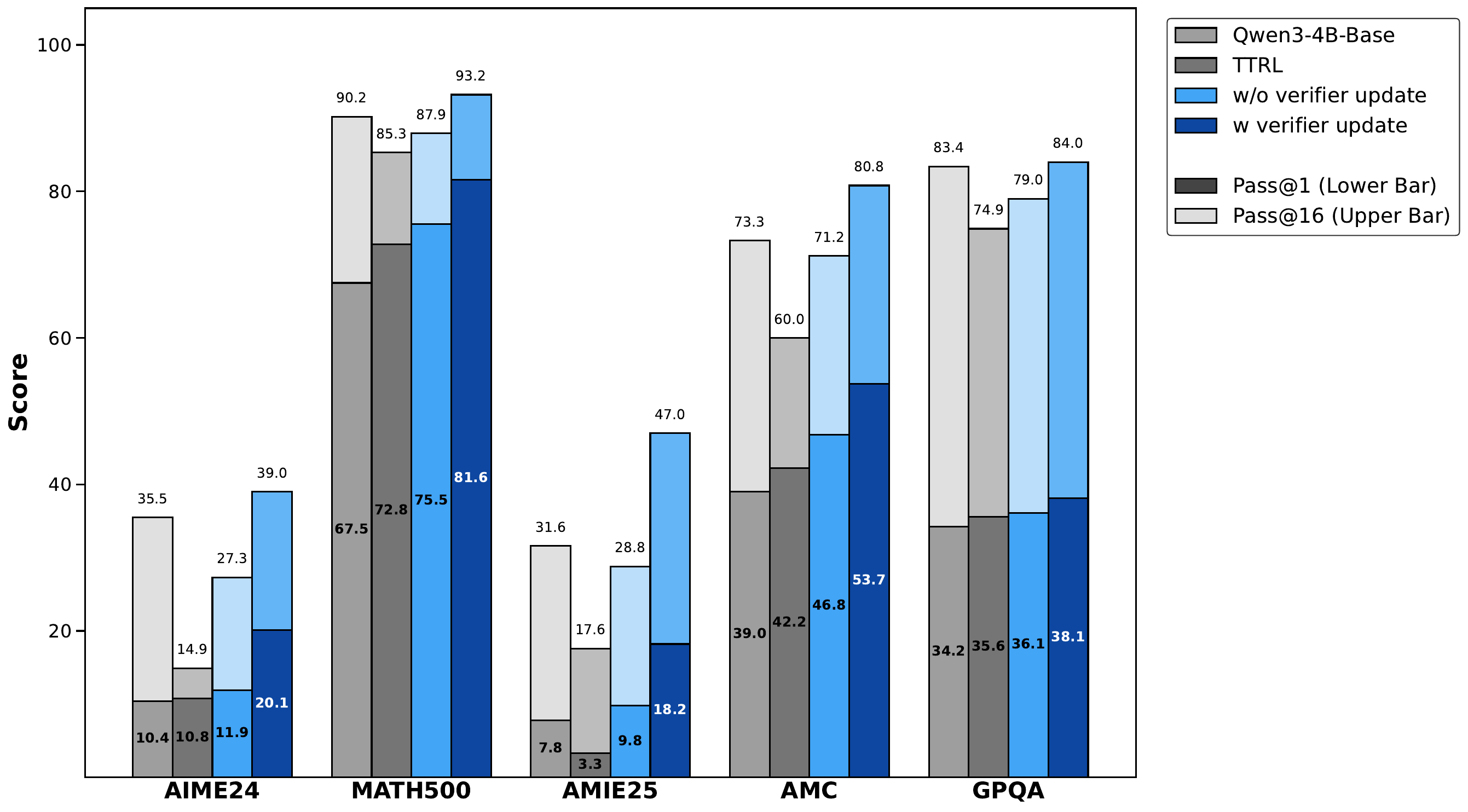}
    \caption{Impact of verifier co-evolution on downstream task performance. Freezing the verifier (w/o verifier update) consistently degrades performance across all five benchmarks, while joint generator-verifier updates (w/ verifier update) achieve the best results.}
    \label{fig:Q2_1}
\end{figure}

As shown in Fig.~\ref{fig:Q2_1}, the absence of the co-evolution mechanism leads to a significant degradation in downstream task performance. Specifically, on the highly challenging AIME25 benchmark, Pass@1 drops from 18.2\% to 9.8\%, while Pass@16 plummets from 47.0\% to 28.8\%. These results intuitively demonstrate that the co-evolution of the generator and verifier is crucial for enhancing the model's exploration capabilities.

\begin{figure}[h]
    \centering
    \includegraphics[width=0.8\linewidth]{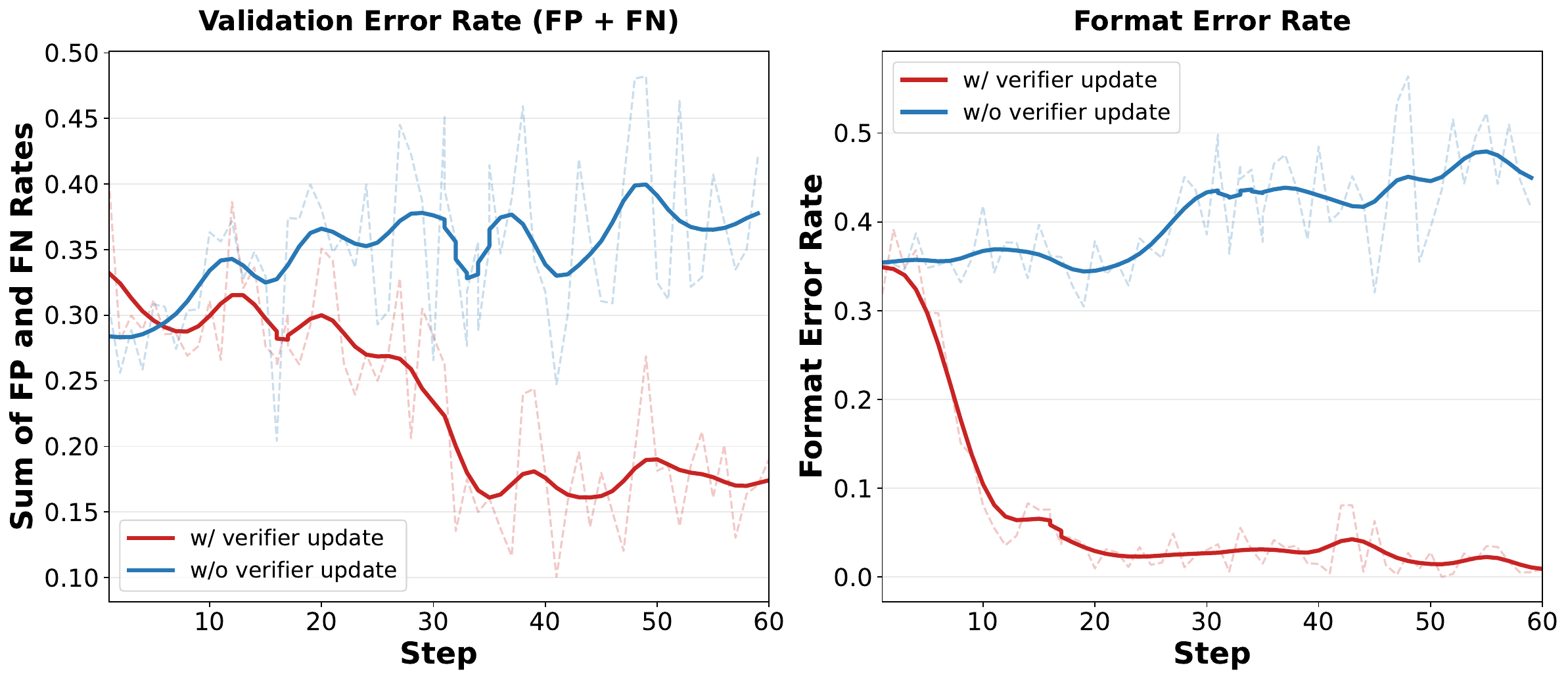} 
    \caption{Detailed internal verification error dynamics. Under joint updates (w/ verifier update), both the validation error rate (FP+FN) and format error rate drop significantly. Conversely, a static verifier (w/o verifier update) suffers from capability mismatch, leading to an escalating validation error rate.}
    \label{fig:appendix_error_rates}
\end{figure}

Regarding the internal verification dynamics, while the main text illustrates the steady climb of the validation correct rate under joint updates (refer to Fig.~\ref{fig:combined_dynamics}, right), we present the corresponding detailed dynamics of the validation error rate and format error rate in Fig.~\ref{fig:appendix_error_rates}. 

Without updating the verifier, the generator's problem-solving ability continues to improve during fine-tuning, yet the static verifier's discriminative upper bound remains locked at the level of the base model. This capability mismatch causes the validation error rate to rise above 0.35. In contrast, when co-updating is enabled, region A continuously provides high-quality positive examples, while region B supplies filtered data samples to assist verifier updates. Consequently, the overall validation error rate and the format error rate drop significantly, ultimately converging to a negligible level.

\subsection{Detailed Impact of Length Diversity Reward}
\label{sec:appendix_diversity}

To further illustrate the impact of the length diversity reward ($R_{div}$) on preventing shortcut learning as discussed in the main text, we compare the dynamic curves of response lengths throughout the training process. 

\begin{figure}[h]
    \centering
    \includegraphics[width=0.75\linewidth]{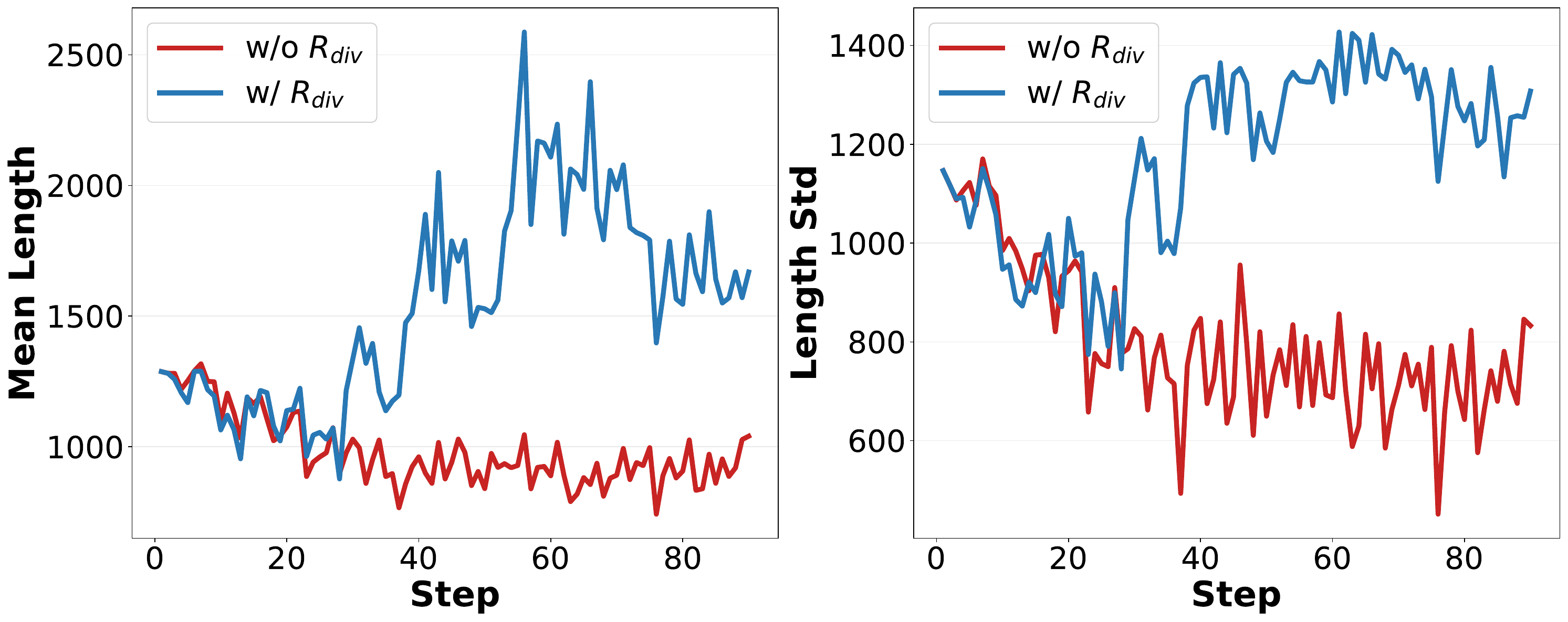}
    \caption{Impact of the length-diversity reward ($R_{div}$) on the mean and standard deviation of response lengths. Without the diversity penalty (red), the standard deviation sharply drops to ~600-800, indicating severe mode collapse and shortcut learning. With $R_{div}$ enabled (blue), the model sustains robust trajectory diversity (Std ~1200-1400) while maintaining correct mathematical intuition.}
    \label{fig:Q3}
\end{figure}

As shown in Fig.~\ref{fig:Q3}, we observe a significant difference when ablating this mechanism. After removing the length diversity reward, not only does the model's average response length rapidly drop to around 1000, but the standard deviation of response length also shrinks severely to the range of 600-800 in the later stages of training. This indicates that the generated responses become highly homogenized. 

In contrast, with the introduction of $R_{div}$, the model's average response length not only avoids shortening but actually grows significantly to the 1500-2500 range. Furthermore, after an initial period of fluctuation, the standard deviation of length rises rapidly and remains robustly at a high level of 1200-1400. This dynamic comparison intuitively reveals that $R_{div}$ provides effective bonuses to correct trajectories that deviate from the mean length, ensuring the model maintains high trajectory diversity.

\subsection{Detailed Evaluation of Scalability Across Model Sizes}
\label{sec:appendix_scalability}

To further elaborate on the scalability discussed in the main text, we conduct cross-scale experiments on the Qwen series of models. 

\begin{figure}[h]
    \centering
    \includegraphics[width=0.9\linewidth]{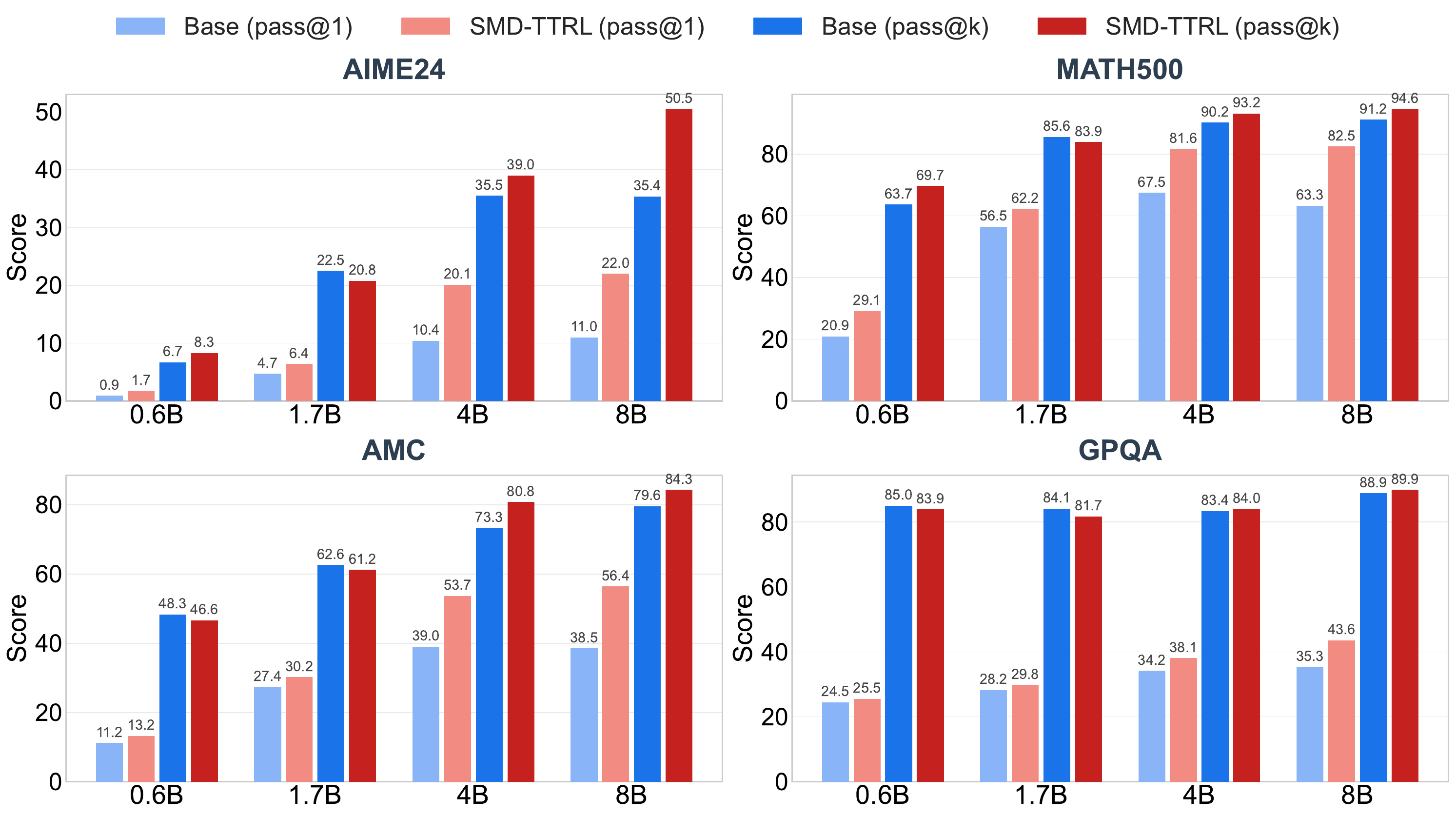}
    \caption{Scalability of TTRL-CoCoV across model sizes. TTRL-CoCoV yields consistent Pass@1 improvements across all model sizes. For Pass@k, the gains become more pronounced as model size increases (4B and 8B), suggesting that the co-evolution mechanism benefits more from larger model capacity. Performance on smaller models (0.6B and 1.7B) is mixed, with improvements on AIME24 and MATH500 but slight declines on AMC and GPQA.}
    \label{fig:Q4_1}
\end{figure}

As shown in Fig.~\ref{fig:Q4_1}, the experimental results indicate that our method consistently outperforms the baselines on core reasoning tasks such as AIME24 and MATH500. More importantly, it exhibits performance gains that increase significantly with the model size. This trend is consistent with the scaling law in reinforcement learning: larger models possess stronger reasoning and verification capabilities, thereby providing more accurate error-filtering signals to our verification mechanism and yielding greater performance improvements.

\subsection{Detailed Evaluation of Generalization Across Model Architectures}
\label{sec:appendix_generalization}

To evaluate the generalization capabilities and the impact of the base model's prior knowledge, we perform cross-model comparisons under similar parameter scales (7B/8B). 

\begin{figure}[tbh]
    \centering
    \includegraphics[width=0.9\linewidth]{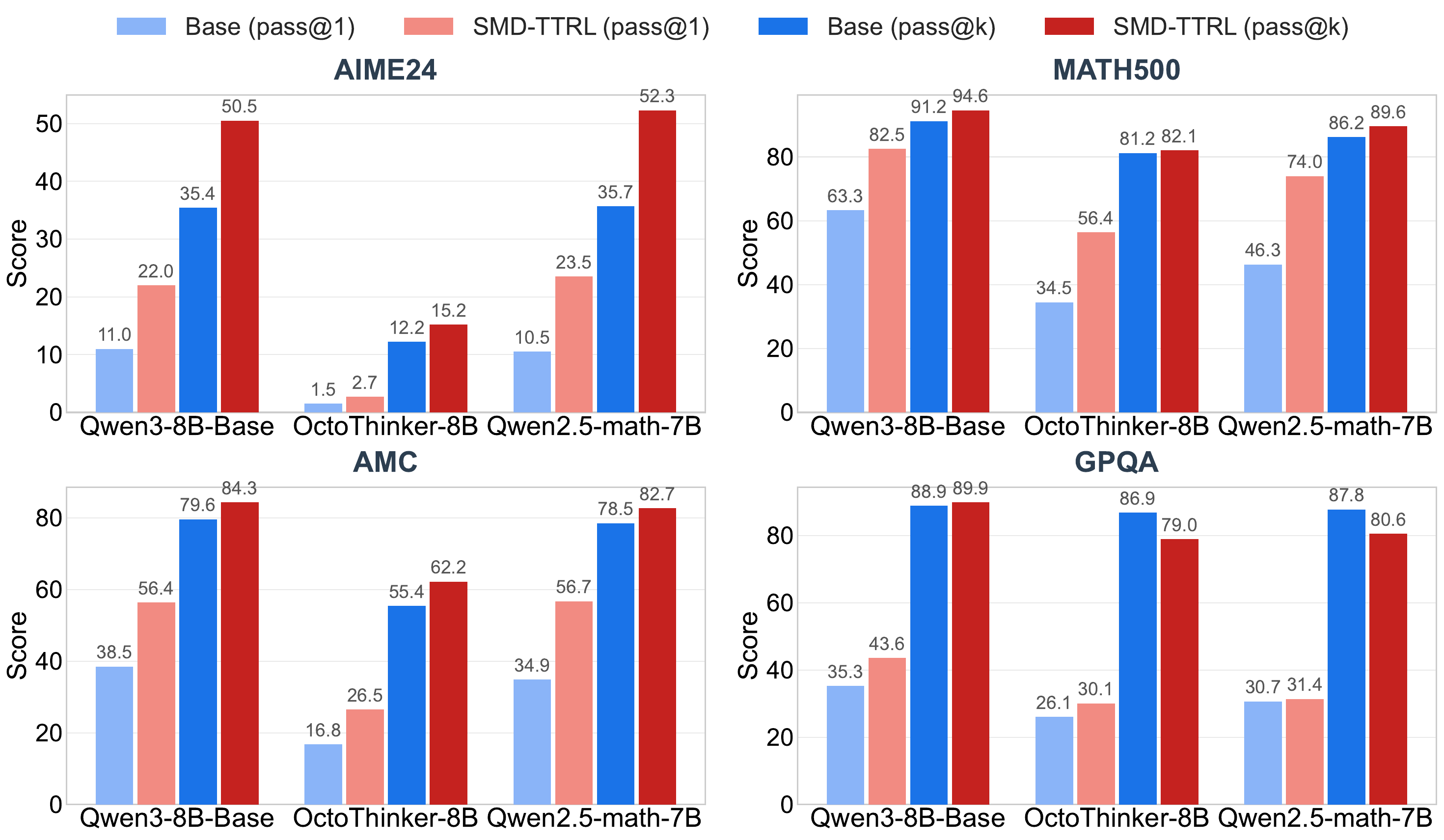}
    \caption{Generalization of TTRL-CoCoV across different models (7B/8B scale). We compare three models: Qwen3-8B-Base, OctoThinker-8B, and Qwen2.5-math-7B on AIME24, MATH500, AMC, and GPQA. Our method consistently improves Pass@1 across all models and benchmarks. For Pass@k, gains are most pronounced on AIME24 and AMC. Despite OctoThinker-8B's weaker base performance, TTRL-CoCoV brings substantial improvements, confirming the method's robustness across diverse architectures.}
    \label{fig:Q4_2}
\end{figure}

As shown in Fig.~\ref{fig:Q4_2}, whether on OctoThinker-8B, a hybrid base model focused on general capabilities, or Qwen2.5-Math-7B, a domain-specific expert model, TTRL-CoCoV achieves stable performance improvements. However, the experiments also reveal a dependency on domain knowledge: our method achieves the most significant gains on math-fine-tuned models when solving complex reasoning tasks, but shows relatively modest improvements on general-domain datasets that cover broad common sense, such as GPQA. This phenomenon indicates that the performance improvement of our method is highly correlated with the base model's original capability in the target domain. 

\newpage
\subsection{Impact of Asymmetric Reward on False Positive Suppression}
\label{sec:appendix_asymmetric}
\begin{wrapfigure}{r}{0.4\textwidth}
    \vspace{-10pt} 
    \centering
    \includegraphics[width=\linewidth]{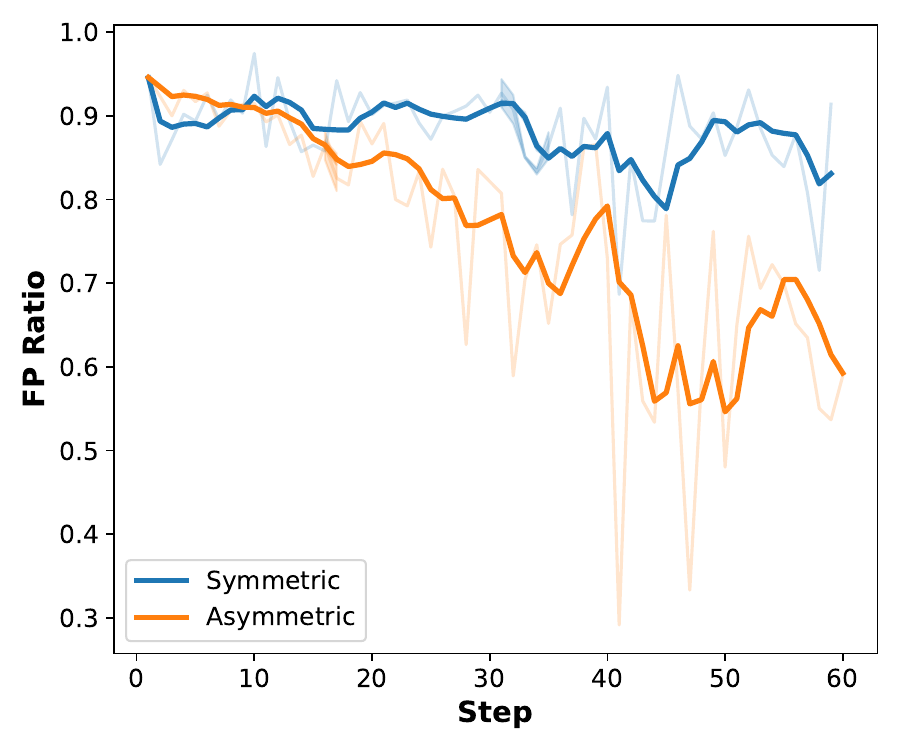}
    \caption{FP ratio dynamics under symmetric vs. asymmetric reward strategies.}
    \label{fig:Q5}
\end{wrapfigure}
To further evaluate the necessity of the asymmetric soft penalty reward matrix discussed in the main text, we compare the verification dynamics against a symmetric reward baseline, focusing specifically on the false positive (FP) ratio.

Empirical results (see Fig.~\ref{fig:Q5}) demonstrate a stark divergence in verifier behavior: under the symmetric setting, the verifier fails to sufficiently discriminate incorrect responses, leading to a persistently elevated false positive ratio that introduces significant noise into the generator's gradient signals. Conversely, the asymmetric strategy, which assigns a higher penalty weight to false positives, imposes a more rigorous screening criterion. Consequently, the false positive ratio is notably reduced to approximately 0.6. This effectively blocks erroneous gradient updates from corrupting the generator.

\subsection{Ablation Study}
To further validate the individual contributions of the proposed components in TTRL-CoCoV, we conduct additional ablation study based on the Section \ref{analysis}, using Qwen3-4B-Base. Table \ref{tab:ablation-study} presents the additional results. 

\textbf{Necessity of the Verification-Guided Evolutionary Paradigm}
We first isolate the fundamental contribution of our dual-role architecture by comparing the w/o $R_{div}$ variant (which retains the internal verifier) against the naive baseline (w/o verifier+$R_{div}$). The dismantling of the verification system leads to a consistent performance deterioration across all reasoning benchmarks. This degradation is particularly pronounced on highly complex datasets where generation uncertainty is pervasive; for instance, Pass@1 on AIME25 drops from 13.5 to 8.5, and AIME24 falls from 13.7 to 11.6. These results empirically substantiate that a generation-only evolutionary process is inherently fragile. Without the verifier acting as an internal safeguard to continuously screen and filter low-confidence samples, the model inevitably falls victim to the consensus trap, internalizing severe pseudo-label noise that fatally destabilizes the policy optimization trajectory.

\textbf{Impact of the Trajectory Diversity Bonus ($R_{div}$)}. The absence of $R_{div}$ triggers a dramatic contraction in the model's generation coverage, severely bottlenecking Pass@16 performance. Specifically, Pass@16 plummets from 47.0 to 32.4 on AIME25, and from 93.2 to 89.8 on MATH500. More critically, this diversity collapse directly impairs the model's peak exploitation capability, evidenced by the corresponding sharp declines in Pass@1 (e.g., an absolute drop of 6.4 on AIME24 and 8.3 on AMC). This confirms our core hypothesis: $R_{div}$ is indispensable for sustaining a broad exploratory space, which provides the essential foundation for discovering novel, robust reasoning trajectories.
\begin{table}[htbp]
  \centering
  \caption{Ablation study results on Qwen3-4B-Base. Results are reported as  Pass@1 / Pass@16 for all datasets.}
  \label{tab:ablation-study}
  \begin{tabular}{lccccc}
    \toprule
    Method & AIME24 & MATH500 & AMIE25 & AMC & GPQA \\
    \midrule
    TTRL-CoCoV & \textbf{20.1} / \textbf{39.0} & \textbf{81.6} / \textbf{93.2} & \textbf{18.2} / \textbf{47.0} & \textbf{53.7} / \textbf{80.8} & \textbf{38.1} /\textbf{ 84.0 }\\
    \midrule
    w/o verifier+$R_{div}$ & 11.6 / 27.0 & 75.6 / 86.7 & 8.5 / 24.6 & 47.1 / 72.1 & 36.1 / 75.0 \\
    w/o $R_{div}$ & 13.7 / 28.6 & 76.9 / 89.8 & 13.5 / 32.4 & 45.4 / 72.2 & 35.3 / 77.1 \\
    \bottomrule
  \end{tabular}
\end{table}

\section{Detailed formulations}

\subsection{Mathematical Formulation of GRPO}
\label{grpo}
In the section \ref{gen_inst}, we briefly introduce Group Relative Policy Optimization (GRPO) \citep{shao2024deepseekmath}. Here, we provide the complete mathematical formulations. For a given input $\boldsymbol{x}$, the policy samples $N$ responses $\{\boldsymbol{y}_i\}_{i=1}^N$. Each response receives a reward $R_i$. The sequence-level advantage $\hat{A}_i$ is estimated by normalizing the rewards within the sampled group:$$\hat{A}_i = \frac{R_i - \text{mean}(\{R_j\}_{j=1}^N)}{\text{std}(\{R_j\}_{j=1}^N)}.$$Using this normalized advantage, GRPO optimizes the policy $\pi_\theta$ via the standard clipped surrogate objective, formulated as:$$\begin{aligned}
\mathcal{J}(\theta) &= \mathbb{E}_{\boldsymbol{x} \sim \mathcal{D}, \boldsymbol{y} \sim \pi_{\theta_{\text{old}}}} \Bigg[ \frac{1}{|\boldsymbol{y}|} \sum_{t=1}^{|\boldsymbol{y}|} \min \Bigg( \frac{\pi_\theta(y_t \mid \boldsymbol{x}, \boldsymbol{y}_{<t})}{\pi_{\theta_{\text{old}}}(y_t \mid \boldsymbol{x}, \boldsymbol{y}_{<t})} \hat{A}_i, \\
&\qquad \text{clip} \left( \frac{\pi_\theta(y_t \mid \boldsymbol{x}, \boldsymbol{y}_{<t})}{\pi_{\theta_{\text{old}}}(y_t \mid \boldsymbol{x}, \boldsymbol{y}_{<t})}, 1 - \epsilon, 1 + \epsilon \right) \hat{A}_i \Bigg) \Bigg]
\end{aligned}$$This formulation allows the model to optimize its reasoning policy efficiently using internal relative rankings without relying on an external, independently trained value network.

\subsection{Analytical Advantage for Pass@k Training}
\label{passk}
In Section \ref{gen_inst}, we refer to the analytical advantage $A^{pass@k}$ derived for Pass@k Training \citep{chen2025pass}. For completeness, we provide the exact closed-form formulations here. Given a group of $N$ responses containing $N_{\text{neg}}$ incorrect samples, the expected group reward is $\bar{R}^{\text{group}} = 1 - \binom{N_{\text{neg}}}{k} / \binom{N}{k}$ and its standard deviation is $\sigma^{\text{group}} = \sqrt{\bar{R}^{\text{group}}(1 - \bar{R}^{\text{group}})}$. The closed-form advantages for positive and negative responses are defined as:
\begin{equation}
\label{eq:pass_at_k_adv}
\hat{A}{\text{pos}} = \frac{1 - \bar{R}^{\text{group}}}{\sigma^{\text{group}}}, \qquad \hat{A}{\text{neg}} = \left(1 - \bar{R}^{\text{group}} - \frac{\binom{N_{\text{neg}}-1}{k-1}}{\binom{N-1}{k-1}}\right) \cdot (\sigma^{\text{group}})^{-1}.
\end{equation}
These state-action advantages serve as the foundational $A^{pass@k}$ terms used during policy optimization.
\subsection{Asymmetric Soft Reward Matrix for Verification}
\label{reward}
In Section 4, we introduce an asymmetric soft reward matrix to shape the verifier into a stringent screening mechanism. Let $R_{\text{second}}^{(j,k)}$ denote the reward for the $k$-th verification trajectory evaluating the $j$-th candidate answer. The exact reward formulation is defined as follows:
$$
R_{\text{second}}^{(j,k)} =
\begin{cases}
-1.0, & \text{Format error}. \\
+1.0, & \text{True Positive / True Negative}. \\
-0.3, & \text{False Negative}. \\
-0.8, & \text{False Positive}.
\end{cases}
$$
This design embodies the principle of being ``lenient to false negatives while strict with false positives''. By assigning a $-0.8$ penalty to False Positives versus a milder $-0.3$ penalty to False Negatives, we explicitly penalize the verifier more heavily for endorsing incorrect reasoning paths than for cautiously rejecting potentially correct ones.

\section{Limitations and Future Work}
\label{limitaions}
\textbf{Ceiling of Co-evolution.} We begin by discussing the performance upper bound of the generator-verifier dynamic. In TTRL-CoCoV, the verifier enhances its discriminative ability using high-confidence consensus from mastered problems, which in turn empowers it to securely mine and learn from filtered samples in uncertain regimes. However, if the underlying language model possesses weak foundational reasoning capabilities, the high-confidence region will be extremely sparse, directly compromising the reliability of the subsequent low-confidence filtering. In such cases, the entire co-evolutionary cycle is starved of high-fidelity training signals, inherently restricting the framework's ability to guide the generator. This observation aligns with the consensus that while label-free RL effectively unlocks a model's existing potential, the ultimate performance ceiling is still fundamentally constrained by the base model's inherent capacity.

\textbf{Limitation of Outcome-Oriented Verification.} We then consider the scope of the verification mechanism. To manage computational overhead and focus on final utility, our verifier primarily evaluates and the extracted candidate answers rather than the full reasoning rollouts. For tasks with deterministic numerical answers, this approach is highly efficient. However, in scenarios where the correctness of the logical process is paramount, such as complex mathematical proofs, outcome-oriented verification might overlook spurious reasoning steps that coincidentally yield the correct answer. To address this limitation, future work could focus on utilizing rollout-level self-evaluation mechanisms to validate the integrity of the intermediate reasoning process.

\end{document}